# Agent2BIM: A Systematic Evaluation of Reality-Capture Configurations and Prior Models for As-Built Modeling

Boyu Wang [a*], Borja García de Soto [a]

[a] S.M.A.R.T. Construction Research Group, Division of Engineering,
New York University Abu Dhabi (NYUAD), Saadiyat Island, UAE
*Corresponding Author

## ABSTRACT

AI agents can generate as-built Building Information Models (BIMs) from reality-capture data, but the evidence required for accurate reconstruction remains unclear. This paper presents Agent2BIM, an agent-based Scan-to-BIM framework, and evaluates seven capture configurations in a research laboratory and a mechanical, electrical, and plumbing (MEP) scene. Inputs include 360° panoramas, smartphone video, smartphone LiDAR, terrestrial laser scanning (TLS), TLS with posed panoramas, and imagery supplemented by prior IFC models. Outputs are compared against TLS-verified reference models for component detection, geometric accuracy, and MEP connectivity. In the laboratory, component recall ranged from 84% to 100%, but only 9% of components reconstructed from panoramas were within 20 cm of their reference positions, compared with 86% using smartphone LiDAR. In the MEP scene, scale-free inputs recovered 11 - 43% of components, versus 75 - 93% for metric inputs. Prior IFC models improved MEP reconstruction but could propagate inaccuracies. The results highlight the complementary roles of visual and metric evidence in agent-based Scan-to-BIM.

**Keywords:** Scan-to-BIM; AI Agents; Large Language Models; Reality Capture; As-built Modeling; Smartphone LiDAR; IFC; MEP.

## 1. INTRODUCTION

Many tasks depend on a reliable model of what was actually built: renovation design, facility management, energy retrofit, and digital twins of existing buildings, as well as acceptance and quality control on construction sites and the coordination of trades during construction [1]. As-built Building Information Models (BIMs) are therefore created from site observations through Scan-to-BIM: the facility or site is captured, the observations are registered, and building elements are recognized and modeled [2]. In current practice, the modeling stage remains predominantly manual. A technician traces walls, openings, furniture, and services from a registered point cloud in authoring software, and this labor usually exceeds the cost of data acquisition itself.

Two decades of research have sought to automate this conversion. Geometry-driven methods extract planes, cylinders, and openings from point clouds; deep learning methods assign semantic and instance labels; and hybrid methods use images to recognize objects that geometry alone cannot distinguish [3-11]. These approaches have made regular architectural and MEP elements increasingly tractable, but three problems still limit their use in practice. First, they depend on data: most require dense, metric point clouds from TLS or survey-grade mobile mapping, and learning-based methods additionally require large annotated datasets. Second, their scope is limited: each pipeline targets a predefined set of element classes and scene types, and transferring it to new components or environments requires redesign or retraining. Third, they lack robustness: their accuracy degrades under clutter, occlusion, and noise, and no automated method yet delivers engineering-grade accuracy reliably enough to replace manual modeling.

Two recent developments bear directly on these problems. First, reality capture has been democratized. Smartphones capture high-resolution images and video, consumer 360° cameras record a room from a few stations, neural scene representations such as 3D Gaussian Splatting (3DGS) turn ordinary video into photorealistic, spatially continuous reconstructions [12-17], and LiDAR-equipped phones produce metric meshes with posed RGB-D frames [18]. Second, large language model (LLM) and vision-language model (VLM) agents [19] can now use software tools, including 3D modeling software operated through code. Agents have been shown to assemble scenes, edit materials, and generate procedural geometry [20-23], and LLMs have begun to produce building models and structured indoor layouts [24,25]. Recent studies have also applied LLM agents to Scan-to-BIM and building reconstruction [26-28]. The Model Context Protocol (MCP) standardizes how agents connect to such tools, so that an agent can execute code in a modeling environment such as Blender, query the scene, render views, and inspect the results [29,30].

These developments change how a model can be produced from site data. Conventional pipelines, whether geometry-driven or learning-based, are fixed sequences of processing steps: the viewpoints, features, and decision rules are set in advance, and the generated model is rarely checked against the data afterwards. This rigidity limits both the range of scenes a pipeline can handle and the accuracy it can achieve. An AI agent works more like a human engineer. It can choose and change its viewpoint, for example by rendering a close-up or a section of a region it finds ambiguous; it can plan a strategy that suits the scene and the data at hand; and it can measure its own result against the evidence and revise the model when the two disagree. The modeling software is only the agent's instrument; what matters is this flexible, closed-loop way of working. Because the agent also absorbs most of the modeling labor, the bottleneck moves to the evidence. The questions that matter become how much metric information the agent needs, which capture modality supplies it most efficiently, and whether a rough model of the fixed building fabric can stand in for dense scanning.

These questions are new and pressing. Agents that can operate 3D modeling software end to end have become available only very recently, for example GPT-6 Astra, released in September 2026, and early reports concentrate on demonstrating what such agents can build. Existing agent-based reconstruction studies typically fix the input modality and target a defined subset of elements, such as walls and spaces or building exteriors. To the authors' knowledge, it has not yet been tested systematically how the choice of capture data affects the semantic and geometric accuracy of agent-generated as-built models at the level of individual components, including furniture, equipment, and MEP networks, and there is no guidance on which capture strategy offers the most practical balance of effort and accuracy in an engineering setting. Studies of LLM-driven 3D generation evaluate visual plausibility or instruction following rather than metric accuracy against a measured reference, and Scan-to-BIM benchmarks evaluate algorithms rather than capture strategies for an agentic modeler. The experiments of this paper illustrate the stakes. From five 360° panoramas and a single door width, an agent recognized 100 of the 103 reference components of a laboratory and produced a convincing model, yet only 9 of those components lay within 20 cm of their true position. From a short smartphone LiDAR scan, the same type of agent placed 89 components within that tolerance. Visual quality and geometric validity can therefore diverge sharply, and the capture configuration determines which one an agent-generated model actually delivers.

This paper conducts a controlled comparison of seven capture configurations for agent-based Scan-to-BIM in two scenes, which yields fourteen agent-generated models. Five configurations address Scan-to-BIM from scratch: two scale-free inputs from consumer cameras (360° photographs and a smartphone video) and three metric inputs (a smartphone LiDAR scan, TLS, and TLS with posed panoramas). Two further configurations address Scan-to-BIM with model updating, combining the scale-free inputs with an existing IFC model of the space. The experiments are deliberately limited to two scenes, a research laboratory and a garage ceiling with a dense multi-system MEP network, so that every configuration can be tested under identical conditions against a carefully verified reference; larger, more numerous, and more complex scenes are left to future work. This paper addresses four research questions:

**RQ1.** How does the capture modality affect the completeness, semantic correctness, and geometric accuracy of agent-generated models?

**RQ2.** To what extent can an existing prior model compensate for the missing metric information in low-cost capture?

**RQ3.** What are the cost–accuracy trade-offs among configurations, and which configurations are the most efficient, the most accurate, and the most practical for typical as-built use cases?

**RQ4.** Which failure modes are characteristic of agent-driven modeling, and which verification mechanisms detect them?

The contributions are fourfold. (1) It systematically reviews the evolution of reality capture, semantic understanding, and Scan-to-BIM methods from survey-grade scanning and rule-based modeling toward foundation models and agents, and identifies the dependencies and gaps that motivate agent-based modeling. (2) It introduces Agent2BIM, an agent-based Scan-to-BIM framework in which an AI agent receives raw capture data and a task prompt and models, renders, measures, and revises the as-built model in a closed loop, using 3D modeling software as a tool. (3) It provides a systematic, reference-based comparison of seven capture configurations in a general indoor scene and an MEP scene, using the same agent and protocol for all inputs, including scale-free versus metric inputs and paired experiments that isolate the contribution of a prior model, in terms of component-level semantic detection, position and geometric accuracy, and network topology; whereas previous agent-based Scan-to-BIM work evaluated a fixed input and a defined set of elements, this comparison isolates the effect of the capture evidence itself. (4) It derives a failure-mode taxonomy and indicative configuration recommendations for practical as-built use cases. The remainder of this paper is organized as follows. Section 2 reviews related work. Section 3 describes the framework and configurations. Section 4 presents the experimental design. Sections 5 and 6 report and discuss the results, and Section 7 concludes the paper.

## 2. LITERATURE REVIEW

Scan-to-BIM can be understood as a chain of three layers. Reality capture determines which geometric and visual evidence is available. Semantic understanding converts that evidence into labelled objects. Scan-to-BIM methods in civil engineering then turn the labelled objects into BIM elements with geometry, type, and relationships. This section reviews the three layers in turn. Fig. 1 summarizes how each layer has evolved, from survey-grade laser scanning and hand-crafted geometric rules toward low-cost capture, foundation models, and agents that act directly on the model. Read across the layers, the trend is a steady relaxation of dependencies: from dense metric data and manually tuned rules, to large labelled datasets, to few or no training samples, and finally to agents that reason about the evidence and build the model themselves.

### 2.1 Reality Capture for As-built Modeling

Accurate spatial information is indispensable for as-built modeling. Traditional measurement with tapes and total stations requires experienced surveyors to record objects one by one, which is time-consuming and error-prone. Reality-capture technologies record the geometry and appearance of an entire scene instead, and they now range from survey-grade laser scanners to consumer smartphones [31,32].

**Terrestrial laser scanning (TLS)**: A tripod-mounted scanner rotates its mirror and body to sweep laser beams across the surroundings and records hundreds of thousands of points per second with millimeter-level accuracy. Distances are measured either from the round-trip time of a pulse (time of flight) or from the phase difference of an amplitude-modulated continuous wave. Scans from multiple stations are registered into a common coordinate system to reduce occlusion. Because of its accuracy and reliability, TLS has become the reference technology for as-built modeling of buildings, infrastructure, and MEP systems [2-4]. Its drawbacks are equipment cost, the time needed to plan and occupy many stations in cluttered spaces, the registration effort, and the absence of semantics beyond intensity and color.

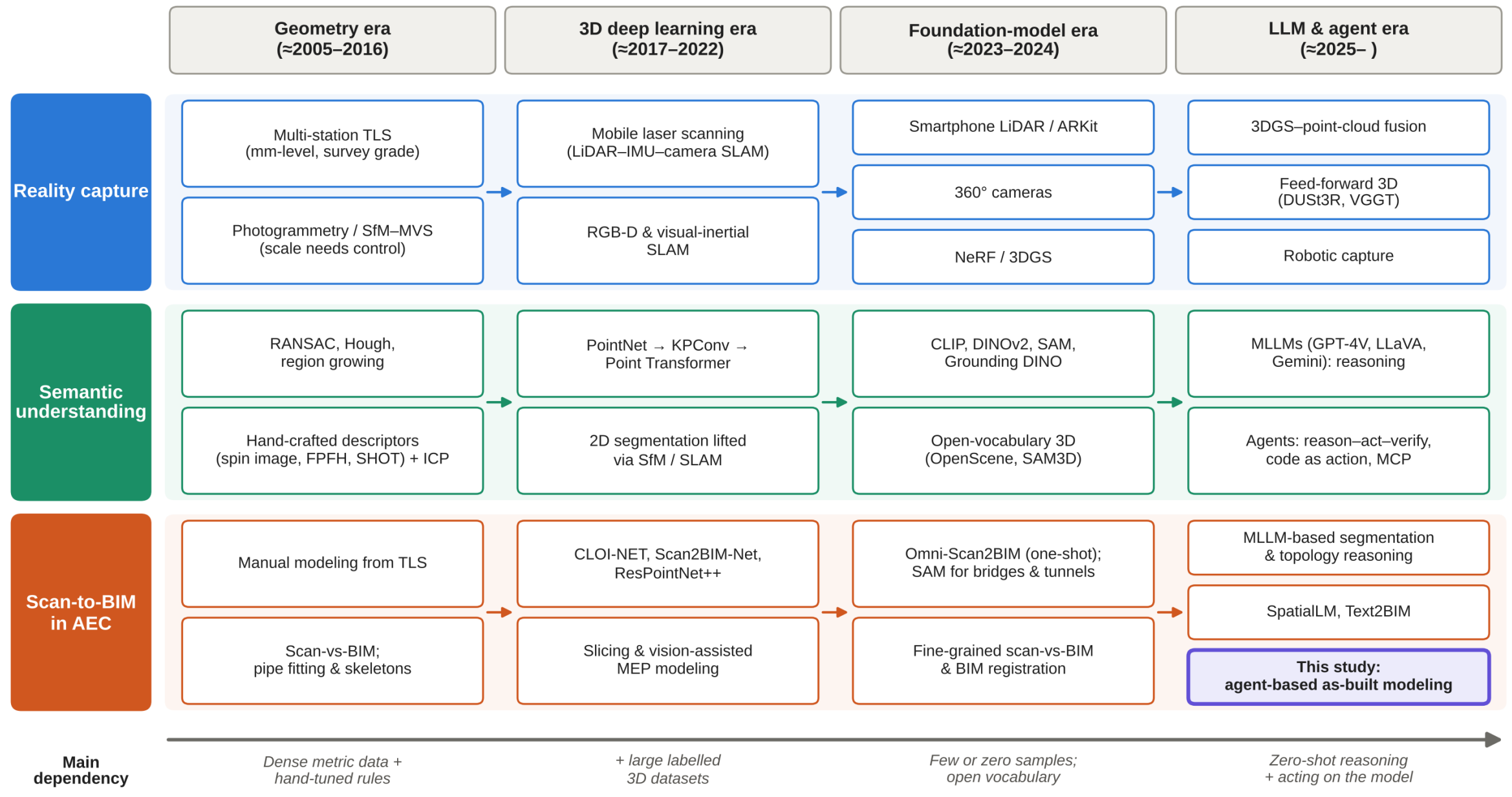

**Fig. 1** Evolution of reality capture, semantic understanding, and Scan-to-BIM methods, and the main dependency of each era.

**Mobile laser scanning (MLS)**: Vehicle-, trolley-, backpack-, and handheld systems combine LiDAR with inertial measurement units and cameras, and register points continuously through simultaneous localization and mapping (SLAM). They capture large indoor areas far faster than TLS, at the cost of centimeter-level accuracy and trajectory drift [33]. Tightly coupled LiDAR–inertial–visual odometry now records posed images alongside the point cloud, so appearance and geometry are acquired together [34], and legged robots carrying such sensors automate capture in cluttered or hazardous sites [35-38].

**Image-based reconstruction**: Photogrammetry recovers camera poses and dense geometry through structure-from-motion (SfM) and multi-view stereo [12]. Unordered site photographs and UAV imagery have been used for progress monitoring, pipe-work tracking, and infrastructure inspection [39-42]. Cameras are cheap and portable, but image-based reconstructions suffer from incomplete feature matching on texture-less or repetitive surfaces, sensitivity to illumination, and a scale that is undefined without control points; their geometric accuracy is generally lower than that of laser scanning [43]. 360° cameras cover a room from a single station and are widely used for site documentation and for localizing images against BIM [44,45], although a few panoramas carry little metric information on their own.

**RGB-D cameras and smartphones**: Depth cameras combined with visual or visual-inertial SLAM build dense, posed maps in real time [46,47], and semantic SLAM attaches labels to these maps [48]. Such maps have been used to locate MEP components approximately before they are refined in LiDAR data [5]. LiDAR-equipped smartphones and tablets bring the same capability to consumer devices: visual-inertial tracking with ARKit provides metric camera poses, and the depth sensor produces textured meshes and posed RGB-D frames within minutes [18]. Their accuracy is at the centimeter level over room-scale distances [49], which is far below TLS but obtained at a small fraction of the cost and effort.

**Neural scene representations**: Point clouds are discrete and often sparse, carry little photometric detail, and show gaps on poorly lit or reflective surfaces. Neural radiance fields (NeRF) instead approximate a continuous volumetric field with a neural network and synthesize photorealistic novel views [13], but the implicit representation is costly to train and render and cannot be edited directly. 3D Gaussian Splatting (3DGS) returns to an explicit, point-based representation: anisotropic Gaussians initialized from SfM points are optimized through differentiable, tile-based rasterization, which retains photorealism while rendering in real time and allowing the scene to be edited [14,50]. Subsequent variants reduce aliasing [51,52], align Gaussians with surfaces for accurate mesh extraction [15,53,54], and organize primitives around structured anchors [55]. Because a 3DGS inherits the scale and pose errors of its reconstruction, it is ill-suited for direct metric analysis; registering it to measured point clouds yields a 3DGS–point-cloud (3DGS-PC) representation that is photorealistic, editable, and metrically accurate [56,57]. In the AEC domain, 3DGS has also been used to digitize historical buildings with language-embedded semantics [58] and reflective glass façades that defeat conventional scanning [59]. Most recently, feed-forward models such as DUSt3R and VGGT estimate dense geometry and camera poses directly from a few images without per-scene optimization [16,17].

Table 1 compares these modalities. The field has moved from a single, expensive, geometry-only reference technology toward many cheap, appearance-rich modalities whose metric quality varies widely. For an automated modeler, the relevant question is therefore no longer which device is most accurate but which combination of modalities supplies sufficient metric and semantic evidence for the intended use. This question has not been examined systematically.

## 2.2 Semantic Understanding: From Geometric Rules to Agents

Semantic understanding assigns object identity to captured geometry. Its methods have passed through six overlapping stages, each of which removed a dependency of the previous one: geometric rules, 3D deep learning, 2D–3D hybrid methods, vision foundation models, multimodal large language models (MLLMs), and agents.

### 2.2.1 Geometric Methods

Early methods exploited the regularity of built objects. Region growing based on smoothness constraints partitions point clouds into surfaces [60], efficient RANSAC detects planes, spheres, cylinders, cones, and tori [61], and Hough voting estimates cylinder directions and radii [62]. For irregular objects, hand-crafted local descriptors such as spin images, FPFH, and SHOT [63-65] and global descriptors such as VFH and ESF [66,67] are matched against templates, and the iterative closest point (ICP) algorithm refines the pose [68]. These methods need no training data and are interpretable, but they depend on carefully tuned parameters, struggle with noise, clutter, and occlusion, and recover shape rather than meaning: a cylinder is detected, not a sprinkler pipe.

### 2.2.2 3D Deep Learning

PointNet first learned features directly from unordered points using shared multilayer perceptrons and symmetric pooling [69]. Subsequent networks captured local structure through hierarchical grouping, point convolutions, dynamic graphs, and efficient sampling [70-74], and transformer-based architectures now lead indoor benchmarks [75,76]. Instance segmentation networks group points into objects [77,78]. Because point-level annotation is expensive, self-supervised pre-training, weakly supervised learning, and few-shot learning have been proposed to reduce label requirements [79-83]. Nevertheless, 3D deep learning models remain specialized: they are trained for a fixed label set on a specific data distribution, generalize poorly to new scenes, and still struggle with small objects and fine-grained types.

**Table 1** Reality-capture modalities for as-built modeling (indicative characteristics).

| Modality | Typical devices | Geometric quality | Metric scale | Appearance | Capture effort and cost | Typical role in Scan-to-BIM |
|---|---|---|---|---|---|---|
| TLS | Tripod scanners (TOF / phase-based) | mm-level | Survey | Point color, intensity | High: many stations, registration; expensive | Reference geometry for modeling and verification |
| MLS / SLAM scanners | Handheld, backpack, trolley, robot-mounted | cm-level; drift-dependent | Survey (SLAM) | Point color; posed images on LiDAR–visual systems | Medium: walk-through capture | Large indoor areas; 3DGS–point-cloud fusion |
| Photogrammetry | DSLR, phone, UAV cameras | cm-level with control; incomplete on texture-less surfaces | Needs control points or a known dimension | Rich | Low cost; processing effort | Progress monitoring; façades; infrastructure |
| 360° panoramas | Consumer 360° cameras | Low from a few stations | Undefined without control | Rich, full coverage per station | Very low | Documentation; image-to-BIM localization |
| RGB-D / smartphone LiDAR | Depth cameras; LiDAR phones and tablets | cm-level, short range | Metric (VIO / ARKit) | Rich (RGB frames) | Very low | Rapid room-scale capture; semantic maps |
| Neural representations | 3DGS / NeRF from posed images | Inherits pose quality; weak on sparse views | Inherits poses | Photorealistic, continuous | Low capture; training time | Appearance-rich rendering for recognition |
| Feed-forward 3D | DUSt3R, VGGT from a few images | Relative geometry | Scale-ambiguous | Rich | Very low | Emerging |

Characteristics are indicative and vary with device, protocol, and scene.

### 2.2.3 2D–3D Hybrid Methods

Image segmentation matured earlier than point-cloud segmentation, and 2D annotation is far cheaper than 3D annotation. Hybrid methods therefore segment images with encoder–decoder networks or transformers [84,85] and lift the results into 3D through SfM or SLAM poses. Two lifting strategies have emerged. Mask projection transfers 2D masks onto the point cloud [86,87], while feature projection fuses image features into a 3D representation that can be queried with text or distilled into a 3D network [88-91]. Hybrid methods combine the semantic richness of images with the metric accuracy of point clouds, but they inherit the limitations of the raw images: targets are hard to isolate in cluttered views, background pixels leak into the 3D labels during projection, processing large image collections is computationally expensive, and occlusions and viewpoint changes produce conflicting labels for the same 3D point. Rendering clean, purpose-selected views from an editable 3D representation such as 3DGS has been proposed to avoid these problems [56].

#### 2.2.4 Vision Foundation Models

Foundation models pre-trained on web-scale data changed the generalization regime. CLIP and ALIGN align images and text in a shared embedding space and support zero-shot classification [92,93]. Self-supervised encoders such as MAE and DINOv2 produce general-purpose visual features that transfer to unseen objects [94,95]. Image-text alignment alone, however, offers little spatial granularity. Grounding models such as GLIP and Grounding DINO therefore formulate detection as phrase grounding: image and text encoders are fused through cross-modal attention, language-guided queries select relevant regions, and a decoder predicts boxes by similarity to the text rather than by a fixed classifier, enabling open-vocabulary detection [96,97]. The Segment Anything Model (SAM) introduced promptable, class-agnostic segmentation [98]; prompted with grounded boxes, it segments arbitrary objects named in text [99], and SAM 3 integrates concept-level understanding directly into segmentation [100]. These models made one-shot and zero-shot recognition practical, removing the need for scene-specific training. They remain perception modules, however: they answer what and where, but they do not reason about context, relations, or design intent.

#### 2.2.5 Multimodal Large Language Models

MLLMs couple a visual encoder with a large language model, through query transformers or simple projection layers, and can follow instructions, describe scenes, and reason over images [19,101-104]. Reasoning-segmentation models interpret complex instructions and output interleaved text, boxes, and masks [105-107], and visual prompting with marks and annotations directs their attention to specific regions [108]. Beyond images, SpatialLM aligns point-cloud tokens with an LLM to output structured indoor layouts [25]. MLLMs add contextual reasoning and verification to recognition, but their outputs are text or labels, their metric reasoning is weak, and they can hallucinate plausible but unobserved content.

#### 2.2.6 LLM Agents

Agents place an LLM in a loop of reasoning and acting: the model plans, calls external tools, observes the results, and revises its plan [109,110]. When the tool is a programming interface, code becomes the action. 3D-GPT and SceneCraft generate Blender scenes procedurally from descriptions [20,21], BlenderAlchemy edits materials and geometry with a vision-language feedback loop [22], and BlenderLLM fine-tunes an LLM to write CAD-oriented Blender scripts [23]. The Model Context Protocol (MCP) standardizes how agents connect to such tools, and open-source servers expose Blender for code execution, scene inspection, and rendering [29,30]. Agents thus move semantic understanding from labelling observations to acting on a model: they can build geometry, render it, measure it, and correct it. Existing agent studies, however, target creative or synthetic content and judge outputs by visual plausibility, not by agreement with a measured as-built condition.

### 2.3 Scan-to-BIM in Civil Engineering

The methods above have been adapted to civil engineering at a delay of a few years, following the same sequence of eras (Fig. 1).

**Architectural and structural elements**: Early Scan-to-BIM research targeted planar and regular elements such as walls, slabs, columns, and openings, typically through plane fitting, region growing, and contextual rules, as reviewed in [1,2]. Region growing with multi-scale features segmented architectural, structural, and MEP surfaces [111], and contextual region growing extracted structural components from construction-site scans [112]. Deep learning was introduced through Scan2BIM-Net, which segments building point clouds for Scan-to-BIM [113], and through instance segmentation trained on BIM-generated synthetic data [114]. Dimensional quality control of precast elements and steel members compares extracted geometry with design values [115,116], and design-intent instances can be matched to point clouds directly [117].

**MEP systems**: MEP systems are dense, interconnected, and frequently deviate from their design, which makes them the most demanding Scan-to-BIM target. Geometry-based methods detect cylindrical pipes by Hough transforms on slices, Gaussian-sphere voting, skeletonization, and curvature analysis [118-121], model equipment from P&ID-derived prior knowledge [122], fit NURBS surfaces [123], and retrieve pipe spools with shape descriptors [124,125]. A slicing-based method automated parametric BIM generation for pipes, ducts, and equipment from TLS data [4,126], and combined TLS and LiDAR point clouds have been used for high-precision MEP reconstruction [127]. Deep learning followed with CLOI-NET and ResPointNet++ for industrial segmentation [128,129] and SE-PseudoGrid for piping-component classification [130], label-efficient variants [131], DeepPipes and PipeNet for pipe reconstruction [132,133], transformer networks and dedicated datasets for building and industrial MEP scenes [134-137], and weakly supervised, synthetic-data, and annotation-free training [138-140]. A deep-feature object verification module was proposed to remove false positives in any Scan-to-BIM pipeline [141]. Hybrid methods used images to localize components and small secondary elements before refinement in LiDAR data [5-7,142]. Vision foundation models then removed scene-specific training: Omni-Scan2BIM recognizes MEP components from a single sample image using DINOv2 and SAM [8], and vision-language reasoning resolves occluded and spatially complex MEP point clouds [10].

**Scan-vs-BIM and model updating**: When a design model exists, as-built data can be compared with it rather than modeled from scratch. Integrating Scan-to-BIM and Scan-vs-BIM enabled construction monitoring of cylindrical MEP components [3], and ICP-based comparison supported compliance checking of pipe spools [143]. Most such methods associate design components with scan data by spatial proximity, which fails when installations deviate substantially. Object-aware Scan-vs-BIM based on fine-grained recognition of family types addresses this limitation and transfers design attributes to the as-built model [9]. Registering local scans to the BIM through modality-invariant visibility maps enables comparison without a global scan [144] and, combined with component association, construction quality inspection [145], and MLLM-based segmentation has recently been combined with instance-level Scan-vs-BIM for construction-site assessment [146]. In computer vision, retrieval-based methods similarly align CAD models to RGB-D scans [147].

**Temporary structures, other infrastructure, and zero-shot segmentation**: Temporary structures have followed the same progression. Scaffold work platforms were first checked with horizontal slicing, Z-coordinate histograms, and RANSAC [148]; RandLA-Net was later used to segment scaffold components for automated compliance checking [149], synthetic data and an upsampling adversarial network addressed data scarcity and sparsity [150], robot dogs automated scaffold scanning [37,38], and UAV images with an object detector assessed the safety of scaffold joints [151]. Most recently, 3DGS and VLMs have been combined to inspect scaffolds by recognizing their components in rendered views [56]. Foundation models have also been adopted for bridges, tunnels, hydropower plants, and transportation infrastructure, typically by applying SAM, CLIP, and open-set detectors to rendered or captured images and fusing the labels back into 3D [152-159]. For construction sites, a human-inspired framework adaptively selects observation viewpoints and uses MLLM reasoning to segment point clouds without any training [11].

**Toward LLM- and agent-driven modeling**: LLMs have begun to generate building content and support engineering decisions. Text2BIM produces building models from natural-language descriptions through a multi-agent framework [24], multi-agent LLM systems combined with graph methods check MEP constructability [160], LLMs combined with BIM-based knowledge graphs support emergency response [161], and a 3D segmentation agent combines 3DGS and VLMs to analyze as-built scenes [57]. Closest to this study, recent work has applied agents to as-built reconstruction: an LLM-enabled multi-agent framework converts indoor point clouds into IFC walls, spaces, and semantic graphs [26], AstraLOD3 reconstructs LOD3 building exteriors from images and sparse structure-from-motion points [27], and Pano2BIM generates approximate IFC models from a floor plan and a single 360° image [28]. Each demonstrates feasibility for one input modality and a defined set of elements. A systematic comparison of different reality-capture configurations and the contribution of prior models to agent-generated BIM accuracy remains unexplored.

**Research Gap**: Table 2 compares the paradigms reviewed in this section. Three gaps follow. First, although recent studies have demonstrated agent-based as-built modeling, systematic evaluations of geometric accuracy across different reality-capture configurations against a common, surveyed reference remain limited. Second, although capture modalities now range from smartphones to TLS, to the authors' knowledge they have not been compared as inputs to the same modeling agent under a controlled protocol, nor quantified whether a sparse prior model can compensate for missing metric information. Third, it has not been assessed jointly whether agent-generated models capture the required components, how accurately they place them, at what level of detail, and at what cost. This paper addresses these gaps.

# 3. AGENT2BIM: AN AGENT-BASED SCAN-TO-BIM FRAMEWORK

## 3.1 Framework Overview

Fig. 2 presents Agent2BIM, the agent-based Scan-to-BIM framework evaluated in this paper, which mirrors how a practitioner would use the agent. (1) The capture data of one configuration, together with the prior model where applicable, are placed in an input folder exactly as exported by the capture devices, and a short task prompt asks the agent to model the as-built scene and names the components of interest, for example doors, tables, chairs, shelves, sockets, and exit signs in the laboratory, or pipes, sprinklers, and cable trays in the MEP scene. (2) An LLM agent connected to 3D modeling software through an MCP server reads the folder, decides how to process each file, writes and executes modeling code, inspects the result by rendering, querying, and measuring, and revises the model until its own checks pass. (3) The deliverable is a semantically organized as-built 3D model, in which every component is a named and classified object, together with the scripts that generated it and an inventory of components and open doubts; for evaluation, all models were exported to IFC4. (4) Each deliverable is registered to a common frame and evaluated component by component against a reference model whose accuracy is verified against the TLS point cloud.

The framework treats the agent as a fixed instrument and the capture configuration as the experimental variable. No configuration-specific pre-processing is performed by the authors. The same agent, tools, prompt structure, and modeling conventions are used across configurations, while the input data and corresponding descriptions vary according to the configuration. For C1 and C2, the agent requests one reference dimension at the beginning of each run, which is provided by the operator. This standardized protocol allows the influence of different capture inputs and prior models on agent-generated BIM performance to be examined.

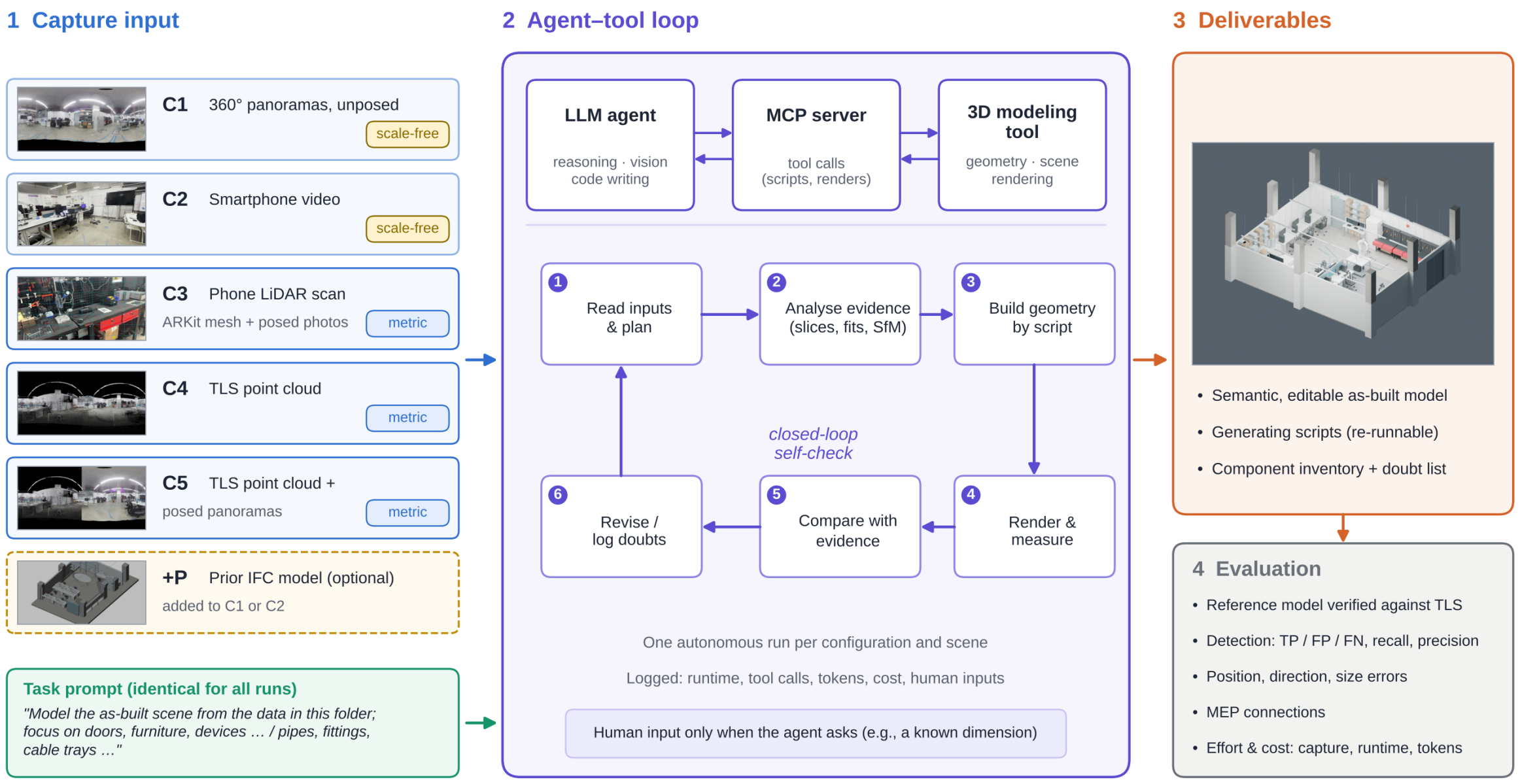

**Fig. 2** Agent2BIM framework and evaluation design. Thumbnails show the S1 inputs, the prior IFC model, and the C1 output.

## 3.2 Capture Configurations

The configurations are organized around two Scan-to-BIM scenarios that occur in practice (Table 3). In the first, Scan-to-BIM from scratch, no previous model exists and the agent must build the model entirely from capture data. In the second, Scan-to-BIM with model updating, an existing model of the space is available and the agent completes and corrects it with new capture data.

**Scan-to-BIM from scratch (C1–C5).** Five configurations span the range from a few photographs to a professional survey. They fall into two groups according to whether the evidence carries metric scale. The scale-free group relies on ordinary imagery: the 360° panoramas recorded at the scanner stations, used without their poses (C1; five panoramas in S1 and seven in S2, 8192 × 4096 px), and a handheld smartphone video of two to three minutes (C2). These inputs define geometry only up to an unknown scale, so the agent requests one approximate reference dimension at the beginning of each C1 and C2 run, which is subsequently provided by the operator. The metric group uses sensors that measure distance. C3 is a scan with a LiDAR-equipped smartphone and an ARKit-based scanning application, which exports the photographs with their camera poses, the depth maps, and a textured mesh. C4 is the registered TLS point cloud, and C5 combines the same point cloud with the station panoramas and their poses. Because C1 and C5 share the same panoramas, and C4 and C5 the same point cloud, the design separates the contribution of metric geometry (C1 versus C5) from that of posed imagery (C4 versus C5).

**Scan-to-BIM with model updating (C1+P and C2+P).** The two scale-free configurations are repeated with an existing IFC model of the space as a prior (P). In S1, the prior is an IFC2X3 model that was drawn manually and roughly from the floor plan. Its walls, columns, slab, and doors are approximately correct, but its 24 furniture items and 24 light fittings were placed without measurement and do not match their actual positions. In S2, the prior is an architectural IFC model with walls, door openings, and slabs but no MEP components. The prior thus supplies a metric frame and the fixed building fabric at no additional capture cost. The metric configurations are not combined with the prior because they already carry metric scale.

**Table 2** Comparison of as-built modeling paradigms.

| Paradigm | Typical input | Metric scale | Semantic coverage | Training data needed | Editable / parametric output | Evaluated against survey reference |
|---|---|---|---|---|---|---|
| Manual Scan-to-BIM [2] | TLS / MLS point cloud | Survey | Unlimited (modeler) | None | Yes | Yes |
| Geometry-based [4,61,118] | TLS point cloud | Survey | Planar and cylindrical elements | None | Partly | Yes |
| 3D deep learning [113,128,129,133] | Point cloud | Survey | Trained classes | Large labelled 3D sets | Rarely | Yes (labels) |
| 2D–3D hybrid [5-7] | Point cloud + images | Survey | Trained 2D classes | Labelled 2D sets | Partly | Yes |
| Vision foundation models [8,10,152] | Point cloud + images | Survey | Open / one-shot | None or one sample | Partly | Yes |
| MLLM reasoning [11,57,146] | Point cloud, images, 3DGS | Survey | Open vocabulary | None | Partly | Yes |
| Scan-vs-BIM / updating [3,9,144] | Point cloud + design BIM | Survey + design | Elements in design model | Varies | Yes (design model) | Yes |
| LLM agents for 3D content and Scan-to-BIM [21,22,24-28] | Text, images, point cloud | Usually none | Open vocabulary | None | Yes (code) | Rarely / no |
| **This paper: Agent2BIM** | 7 capture configurations (Table 3) | Varies by configuration | Open vocabulary | None | Yes (code) | Yes |

The design yields seven configurations per scene and fourteen agent runs in total, and isolates four effects: the effect of metric scale (C1–C2 versus C3–C5), the effect of capture device among metric inputs (C3 versus C4), the effect of posed imagery added to a point cloud (C4 versus C5), and the effect of a prior model (Cx versus Cx+P).

### 3.3 Input Folder and Task Prompt

The input folder contains only the outputs that the capture devices and their standard software produce (Table 3). Device-native processing, such as the registration of the TLS stations and the export of the smartphone scan, is performed beforehand; everything else is left to the agent. The agent decides itself how to read each file type, for example whether to extract frames from a video and run structure from motion, back-project depth maps, slice a point cloud, or project the point cloud into the panoramas, and it writes the code to do so. This design tests the modeling capability of the agent end to end rather than the quality of a hand-crafted pre-processing pipeline.

The task prompt (Appendix A) is short and written as a practitioner would write it. It names the input data and the target components of the scene (Table 5), asks the agent to make the quantity, position, and dimensions of all components accurate, and instructs it not to retain the original capture geometry in the final model, so that the deliverable consists only of modeled components. For the scale-free configurations (C1 and C2), the agent requests one approximate reference dimension immediately after receiving the initial task prompt. This measurement is provided by the operator as the only source of metric scale and recorded as a human input: the width of the double-door opening (1.70–1.80 m) in S1; the width of the glass entrance door (about 1.9 m) for C1 in S2; and the length of a straight pipe segment above a door (about 0.95 m) for C2 in S2. In the model-updating configurations, the prompt states that the IFC file is an existing model of the space.

**Table 3** Capture configurations evaluated in this study.

| ID | Scenario and group | Input data | Device | Metric scale | On-site capture (min) | Files in the input folder |
|---|---|---|---|---|---|---|
| C1 | From scratch; scale-free | 360° panoramas without poses | Panoramic camera of the Leica BLK360 | No (one reference dimension) | ≈ 1 [a] | 5 (S1) / 7 (S2) equirectangular images, 8192 × 4096 px (.jpg) |
| C2 | | Smartphone video | iPhone 14 Pro | No (one reference dimension) | 2 / 3 [b] | Video file, 1920 × 1080 px, 120 s (S1) / 179 s (S2) (.mp4) |
| C3 | From scratch; metric | Smartphone LiDAR scan | iPhone 14 Pro (LiDAR), ARKit-based scanning app | Yes (LiDAR depth, ARKit poses) | 2–3 [c] | Photographs with camera poses, depth maps, textured mesh (.obj) |
| C4 | | TLS point cloud | Leica BLK360 | Yes (survey) | 20–30 [c] | Registered colored point cloud (.las, RGB): 7.26 M points (S1), 14.57 M points (S2) |
| C5 | | TLS + posed 360° panoramas | Leica BLK360 | Yes (survey) | 20–30 [c] | As C4, plus the C1 panoramas with position and orientation files |
| C1+P | Model updating | C1 + prior IFC model | As C1 | Yes (prior) | ≈ 1 [a] | As C1, plus IFC model |
| C2+P | | C2 + prior IFC model | As C2 | Yes (prior) | 2 / 3 [b] | As C2, plus IFC model |

P = prior model: a roughly drawn IFC2X3 model with architecture, furniture, and lighting (S1), and an architectural IFC model without MEP components (S2). On-site capture times: [a] estimated, not measured: the panoramas used here were recorded by the TLS during its survey, and about 1 min is the expected time for a consumer 360° camera (e.g., Insta360) at the same stations; [b] measured from the video durations (S1 / S2); [c] approximate times recorded on site. C1+P and C2+P use the same captures as C1 and C2.

## 3.4 Agent Protocol and Tool Interface

The agent operates the modeling software through an MCP server and works almost entirely through generated Python code: it reads and analyzes the input files with standard libraries, builds the geometry through the Blender Python interface, renders views of its own model, and reads the scene back to check it. Where needed, it installs or calls further tools itself, for example an open-source structure-from-motion package for video frames [12] and an IFC toolkit for reading the prior models. Each run starts from an empty scene and the initial task prompt. For C1 and C2, the operator responds to an initial agent request for a reference dimension. No manual correction of the generated models is performed during the runs. Each run ends with the model file, the generating scripts, a component inventory, and a written summary of the modeling approach and its open issues, which is analyzed in Section 5.1. No limit was set on run time, tool calls, or tokens. Table 4 summarizes the settings.

## 3.5 Modeling Target and Element Taxonomy

The modeling target is defined by the component taxonomy in Table 5. In S1, the classes span architectural elements, furniture, displays and irregularly shaped robots such as a robot arm, a mobile robot, and quadruped robots, and small wall- or ceiling-mounted lighting, electrical, and fire-safety components such as sockets, fire extinguishers, and exit signs. Robots have no standard parametric form and are rarely covered by Scan-to-BIM methods, so they test how the agent represents complex, non-building objects; small wall-mounted components are known to be difficult for point-cloud methods because they contain few points [5,6,142]. In S2, the target is the MEP network under the ceiling: fire-protection pipes, other service pipes, sprinklers, cable trays, and fittings. Elbows and tees that are linked through short pipe pieces are counted as one fitting group, couplings and wall penetrations are not counted, and independent pipe pieces shorter than 0.1 m are excluded.

**Table 4** Agent and software configuration.

| Item | Setting |
|---|---|
| Agent / LLM | GPT-6 Astra (released September 2026), high reasoning effort |
| Agent access | The input folder, a Python environment with shell access on the workstation (the agent could install open-source packages), and the 3D modeling software through MCP |
| Modeling software | Blender 5.2, scripted through its Python API and connected to the agent through an MCP server |
| Tools used by the agent | Python scientific stack for point-cloud, mesh, and image analysis (array computing, robust fitting, spatial indexing, image I/O); video frame extraction and open-source structure from motion (C2, C2+P); IFC parsing (C1+P, C2+P); rendering of check views in the modeling software (all) |
| Task prompt | Standardized natural-language prompt per scene (Appendix A), adapted to each configuration's input data; specifies target components and accuracy requirements |
| Runs | One run per configuration and scene (14 runs) |
| Budget per run | No limit on run time, tool calls, or tokens |
| Human involvement | For C1 and C2, the operator provides one reference dimension in response to an initial agent request. No further human guidance or manual model correction during the runs |
| Outputs | Model file, generating scripts, component inventory, modeling summary; models exported to IFC4 for evaluation |
| Hardware | Laptop with Intel Core Ultra 9 275HX CPU, NVIDIA GeForce RTX 5080 Laptop GPU, and 64 GB RAM |

**Table 5** Component taxonomy and number of reference components.

| Scene and category | Component classes (reference count) | Total |
|---|---|---|
| S1 – Architecture | Walls (4), columns (4), floor (1), doors (2) | 11 |
| S1 – Furniture and fixtures | Tables (14), chairs (13), shelves (2), cabinets (11), whiteboards (3), ladders (1), trolley (1) | 45 |
| S1 – Displays and robots | Monitors (16), robot arms (1), mobile robots (1), quadruped robots (2) | 20 |
| S1 – Lighting, electrical, and fire safety | Lighting (12), wall sockets (6), fire extinguishers (3), other fire-safety equipment (4), exit signs (2) | 27 |
| S2 – MEP | Fire-protection pipe sections (27), other service pipe sections (27), sprinklers (9), cable-tray sections (8), fitting groups (36) | 107 |

S1 components are counted as physical assemblies (for example, a double door is one door and a continuous shelving run one shelf). Desk sockets, power strips, task lamps, and objects outside the common laboratory area are excluded. S2 pipe and tray sections are counted between fittings within the evaluated region.

# 4. EXPERIMENTAL DESIGN

## 4.1 Test Scenes

Scene S1 is the S.M.A.R.T. Lab, a research laboratory at New York University Abu Dhabi (Fig. 3a–d). Its floor envelope in the reference model is 10.15 m × 7.92 m. The space combines regular architectural geometry with 103 target components: dense workstations and storage, a robot arm, a mobile robot, two quadruped robots, and small wall-mounted electrical and safety components. It is representative of the cluttered interiors that conventional automated Scan-to-BIM handles poorly.

Scene S2 is a corridor and the adjacent parking area on a basement parking level of a residential tower (Fig. 3e–h). Under its ceiling run several parallel and crossing systems: red fire-protection mains and branches with upright sprinklers, chilled-water and condensate pipes, insulated service pipes, and cable trays and trunking. The evaluated region measures about 10.5 m × 12.5 m in plan and contains 107 reference components with about 119 m of pipe and 60 m of cable tray. Dense, parallel runs, small fittings, and crossings make the scene a test of whether the agent can separate and connect systems rather than only recognize individual components. All seven configurations are tested in both scenes.

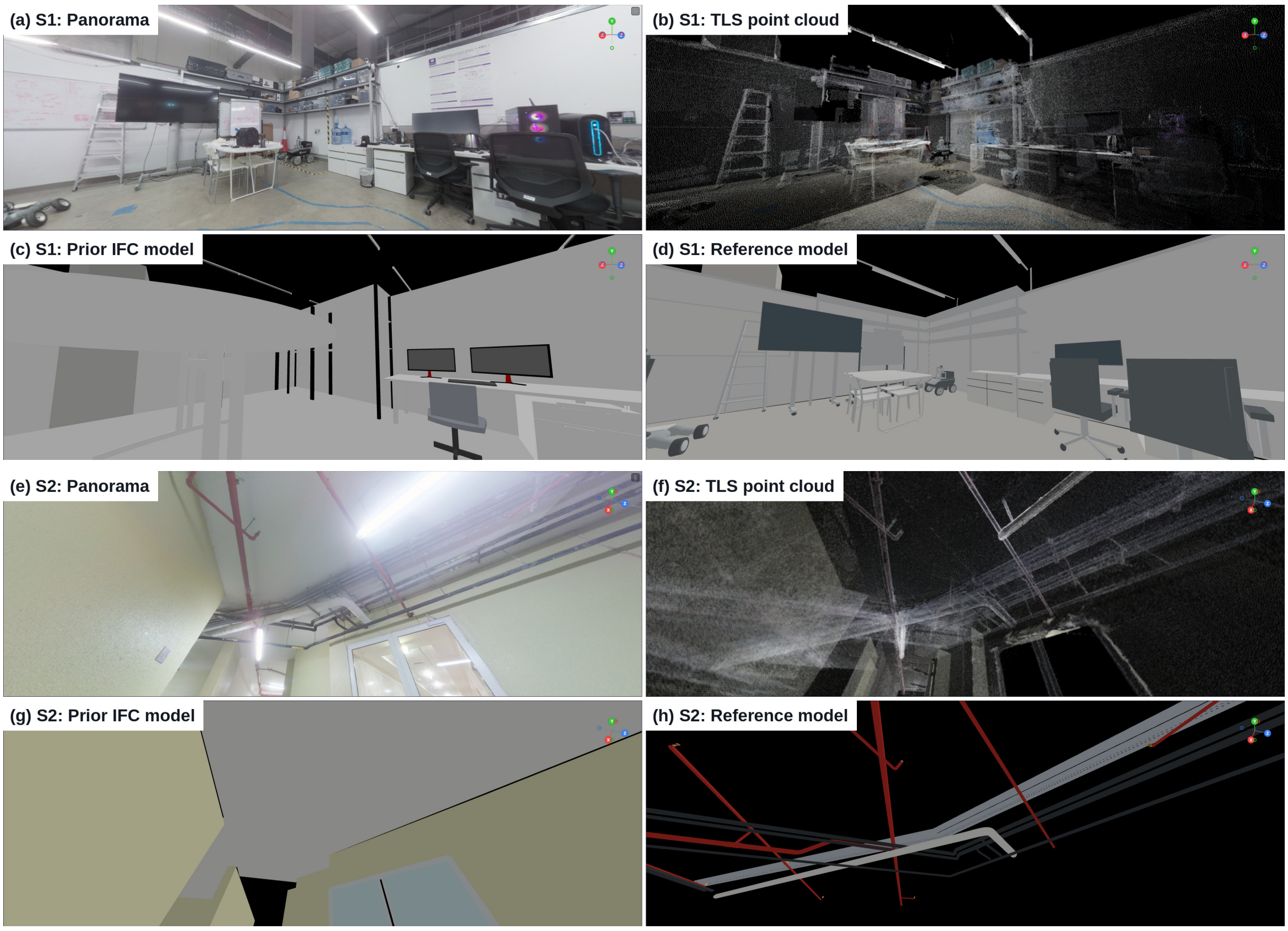

**Fig. 3** Test scenes seen from one viewpoint in a common frame: (a)–(d) S1 and (e)–(h) S2, each showing a view of a station panorama, the TLS point cloud, the prior IFC model, and the reference model.

## 4.2 Data Acquisition

For each scene, the TLS survey, the station panoramas, the smartphone video, and the smartphone LiDAR scan were captured on the same day with the scene left unchanged. The TLS survey used a Leica BLK360, whose integrated camera also recorded the station panoramas; the video and the LiDAR scan were recorded with an iPhone 14 Pro. The S1 point cloud contains 7,259,622 points and the S2 point cloud 14,569,710 points; both are registered in the scanner software and carry RGB color. The panoramas are the 8192 × 4096 px equirectangular images recorded at the five (S1) and seven (S2) scanner stations, with their positions and orientations in the scan frame. The video files given to the agents are 1920 × 1080 px (HEVC) and last 120 s (S1) and 179 s (S2). Each smartphone LiDAR scan took about 2–3 min; in S1 it comprises 339 photographs and the corresponding camera poses, depth maps, and textured meshes with up to 476,156 vertices. Table 3 lists the approximate on-site capture time of each configuration.

## 4.3 Reference Model and Its Verification

The reference ("ground truth") for each scene is a component-level as-built model. It was produced from the TLS point cloud and the posed panoramas, starting from the C5 output, and was refined through multiple rounds of manual inspection against the point cloud and the photographs and prompting-based correction. Each reference component carries its class and, in S2, its system. Although the reference models were refined through manual inspection and verified against the TLS point clouds, their initial development from the C5 outputs introduces a potential bias when evaluating this configuration. Consequently, the geometric agreement reported for C5 should be interpreted with caution, as it does not represent a fully independent evaluation. Nevertheless, the reference models provide a common, TLS-verified basis for assessing the performance of the different capture configurations, subject to the limitations noted.

To verify the geometric accuracy of the reference models, each was sampled into about one million points and compared with its TLS point cloud in point-cloud processing software [162] by computing, for every sampled point, the distance to the nearest scanned point. Fig. 4 overlays each reference model on its point cloud and colors the sampled model points by this distance; Table 6 and Fig. 5 summarize the distributions. In S2, the reference agrees closely with the scan: the median distance is 0.8 cm, 94.1% of the sampled points lie within 3 cm (Fig. 4d).

In S1, parts of the modeled surfaces were not scanned completely: the bases and outer faces of the columns, the faces of furniture against walls or the floor, and the black screens of monitors, which return few laser points (gray areas in Fig. 4b). Model points in such regions have no scanned counterpart, so points farther than 25 cm from the cloud (5.2% of the samples) were excluded from the S1 statistics. For the remaining 948,376 points, the median distance is 3.8 cm, 56.0% lie within 5 cm, and 83.1% within 10 cm. Visible surfaces agree closely; on the upper face of the floor slab, for example, the median distance is 0.5–0.9 cm. The remaining spread stems mainly from hidden faces, such as the underside of the slab and the back faces of walls, whose distance to the nearest scanned point equals roughly the thickness of the element. This verification quantifies the geometric agreement between the reference models and the observed TLS surfaces, although it does not independently establish the accuracy of components or surfaces that were not adequately captured.

## 4.4 Registration

All models are compared in the frame of the reference model, which is the TLS frame. The C4 and C5 models were built in that frame, and so were the model-updating configurations in S2, whose prior IFC model shares it. All other models were registered by a rigid transformation without scaling, so that scale errors remain visible. In S1, a planar rigid transformation was estimated from the four wall midpoints and the two doorway centers, and the floor surface fixed the vertical datum. In S2, where the architecture is sparse, the yaw and translation were estimated from the long fire-protection pipe centerlines. All registered models were exported to IFC4, and the exported files were validated against the IFC schema [163].

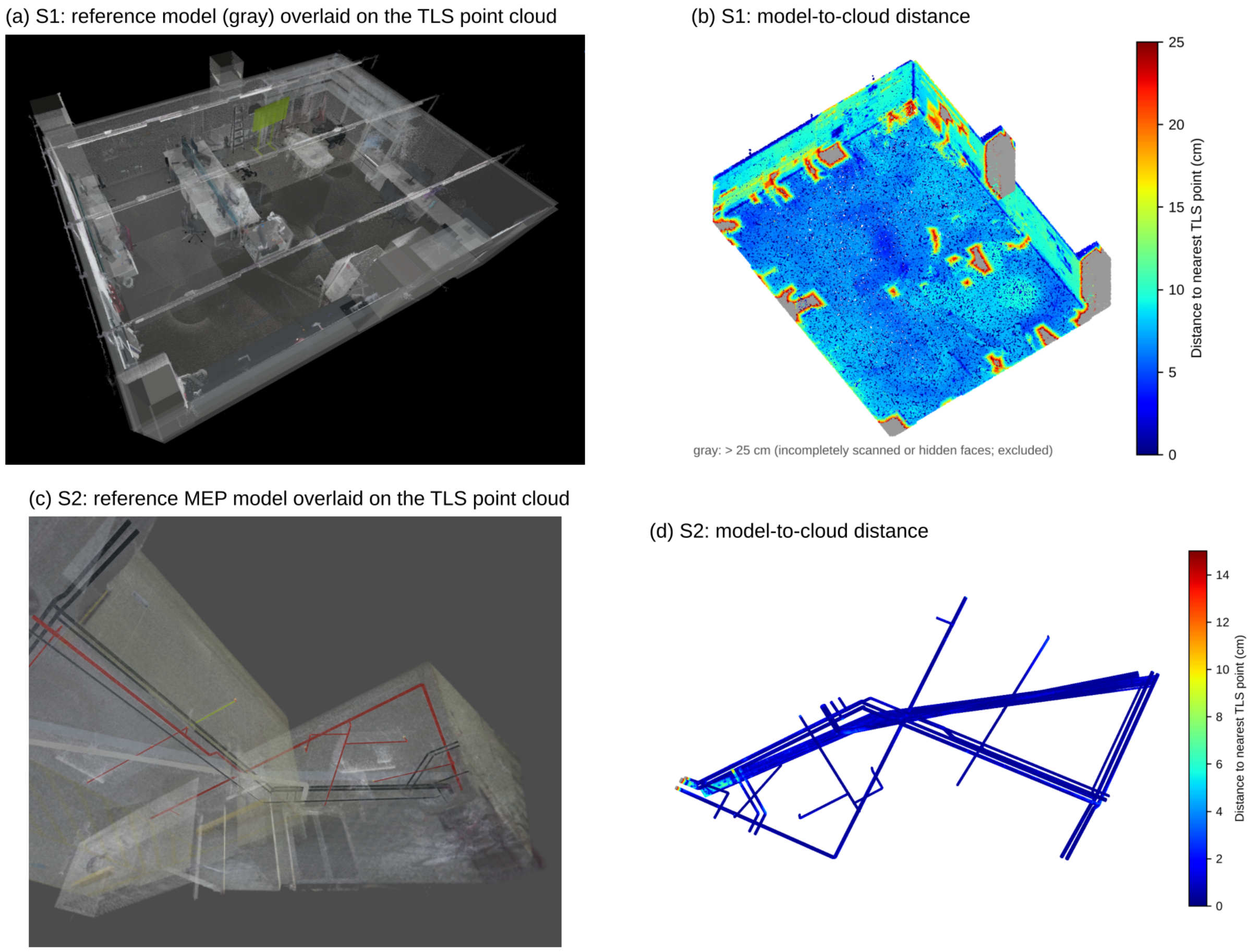

**Fig. 4** Verification of the reference models: (a) S1 reference model overlaid on the TLS point cloud; (b) S1 sampled model points colored by the distance to the nearest TLS point (gray: more than 25 cm, excluded); (c) S2 reference MEP model overlaid on the TLS point cloud; (d) S2 sampled model points colored by the same distance.

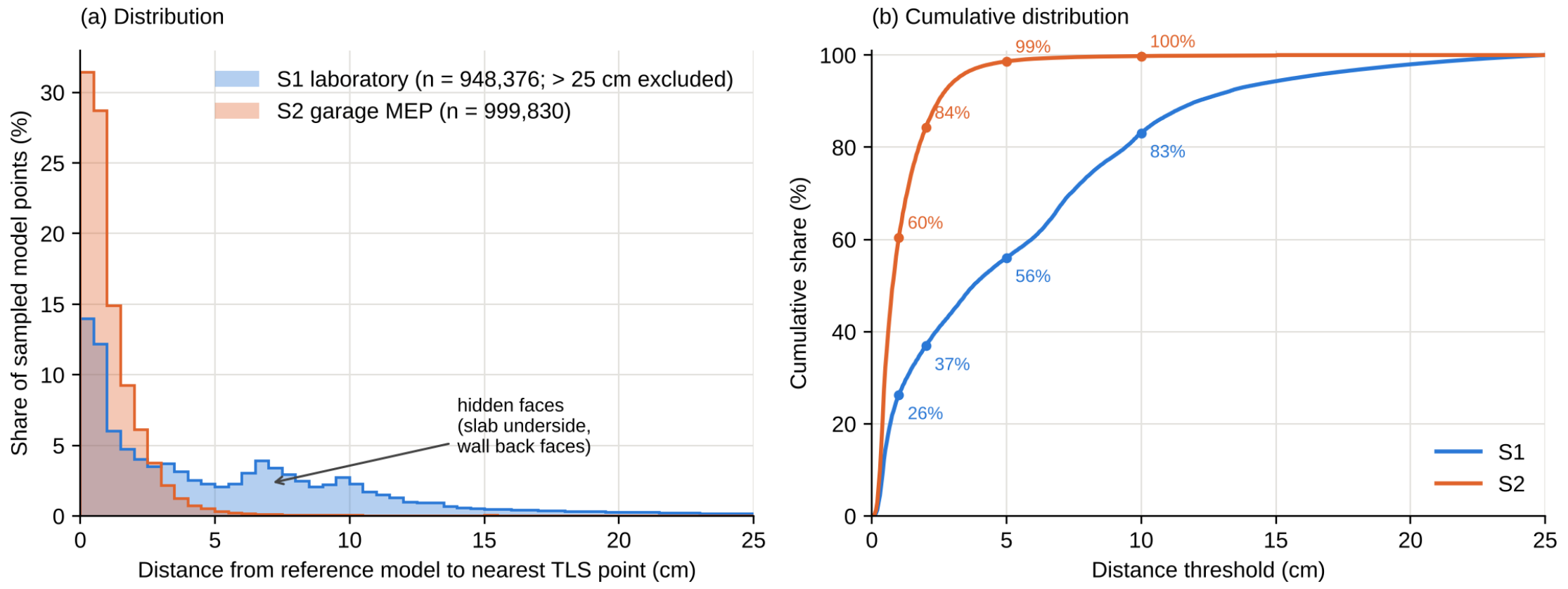

**Fig. 5** Distance from the reference models to the nearest TLS point: (a) distribution of sampled model points; (b) cumulative distribution. S1 excludes points farther than 25 cm.

## 4.5 Evaluation Metrics

The evaluation asks whether each component was modeled, whether it is in the right place, and, for the MEP network, whether its geometry and connections are right (Table 7). The matching rules differ between the scenes because the components differ: S1 contains discrete objects, whereas S2 contains linear components whose identity depends on the network.

**Table 6** Verification of the reference models: distance from sampled model points to the nearest TLS point.

| Scene | Points used | Excluded > 25 cm (%) | Median (cm) | Mean (cm) | ≤ 1 cm (%) | ≤ 2 cm (%) | ≤ 5 cm (%) | ≤ 10 cm (%) |
|---|---|---|---|---|---|---|---|---|
| S1 | 948,376 | 5.2 | 3.8 | 5.4 | 26.3 | 37.0 | 56.0 | 83.1 |
| S2 | 999,830 | 0.0 | 0.8 | 1.2 | 60.4 | 84.3 | 98.6 | 99.8 |

Nearest-point distances from about one million sampled model points per scene (search radius 0.5 m). S1: points farther than 25 cm, which lie in incompletely scanned regions (column bases and outer faces, black screens), are excluded; the remaining distances still include hidden faces such as the slab underside.

**Detection in S1.** Within each class, reference components and model components are paired one to one by minimum-cost assignment on their centers, without a distance limit, so that a component that is recognized but misplaced is still counted as detected and its position error is measured. Unpaired reference components are false negatives (FN) and unpaired model components false positives (FP). Precision is TP / (TP + FP), recall TP / (TP + FN), and F1 = 2TP / (2TP + FP + FN). Detection in S1 therefore measures recognition, that is, whether a component of the right class exists in the model, and not placement: a component can count as detected even if it lies more than 1 m from its true position. Detection and geometric accuracy are thus reported separately and must be read together. To relate detection to engineering tolerance, the recall within 20 cm, that is, the share of reference components detected with a position error of at most 20 cm, is reported in addition. In S1, the floor slab is excluded from all position statistics, because its center depends on the modeled extent of the slab rather than on where a component is placed; it still counts toward the detection metrics.

**Position in S1.** The position error of a true positive is the horizontal distance between the centers of the reference and model bounding boxes after registration; for wall-mounted lighting, electrical, fire-safety, and evacuation equipment, whose height matters, the 3D distance is used. For architectural components, the length, width, and height of the horizontal minimum-area rectangle and the vertical extent are also compared. Orientation is not evaluated in S1.

**Detection in S2.** Pipe and tray sections are matched within the same class by a global one-to-one route allocation that uses centerline distance (seed tolerance 250 mm), direction (seed tolerance 15°), and overlap (at least 50% of the reference length or of the shorter section), supported by evidence from connected neighbors, so that parallel pipes keep their identity. Fitting groups are matched only if at least two of their main-pipe anchors are matched. The seed tolerances only establish correspondences; they are not used to reject matched components for inaccuracy, and a sensitivity analysis varies them (Section 5.3).

**Geometry and topology in S2.** For true positives, the position error is the mean distance between the matched centerlines, the direction error is the angle between their axes, the diameter error is the relative difference of outer diameters, and the length error is the difference of axial lengths. The connection recall is the share of reference connections between components that are recovered between the matched model components.

**Table 7** Evaluation metrics.

| Metric | S1 laboratory | S2 MEP network |
|---|---|---|
| Correspondence | One-to-one assignment within class on box centers, no distance limit | Global one-to-one route allocation within class: ≤ 250 mm and ≤ 15° as seeds, ≥ 50% overlap, neighbor evidence; fitting groups need two matched main-pipe anchors |
| TP, FN, FP; precision, recall, F1 | Per class, category, and scene | Per category and scene |
| Tolerance-based recall | Share of reference components detected within 20 cm | — |
| Position error of TP | Horizontal distance of box centers (3D for wall-mounted devices) | Mean centerline distance |
| Other geometry of TP | Length, width, height of architectural components | Direction (°), outer diameter (%), axial length (mm) |
| Topology | — | Recall of reference connections between components |
| Efficiency and cost | Capture time, agent runtime, tool calls, tokens, operator inputs | As S1 |

# 5. RESULTS

## 5.1 Modeling Strategies of the Agents

The modeling strategies were not prescribed: the prompt only named the input and the targets, and each agent chose how to read its data. They are therefore reported here as observed behavior, reconstructed from the code, intermediate images, and written summaries of the fourteen runs. All runs produced complete, editable models, and every agent followed the same overall loop: it inspected the input, converted it into views it could interpret, identified and de-duplicated components across views, fixed a coordinate frame and a scale, generated parametric geometry with scripts, rendered its own model to compare it with the input, and wrote an inventory and a list of open issues. The decisive differences lay in how each agent turned its input into metric positions (Table 8, Fig. 6).

With 360° panoramas (C1), the agents re-projected each panorama into perspective views (in S1, 40 views from five panoramas; Fig. 6a), identified components across views, and fixed the scale from the reference door width by intersecting viewing rays with an assumed level floor. In S2, the agent first tried feature matching and camera-pose recovery, but the white walls and repetitive floor and pipe patterns made the automatic solution unstable, and it fell back to manual constraints and two-view triangulation of individual sprinklers. With the smartphone video (C2), both agents extracted frames and ran structure from motion (Fig. 6c, d). In S1, all 240 extracted frames were registered with a median reprojection error of 0.74 px. In S2, the agent extracted 358 frames (2 frames/s); the low-texture ceiling split the reconstruction into five separate fragments, of which the agent aligned four through shared features, so that 308 of the 358 frames were registered in one frame and the remaining fragment was used only for identification. With the smartphone LiDAR scan (C3), the agents sliced the mesh by height (Fig. 6e), back-projected selected image points through the depth maps and camera poses, intersected viewing rays from several photographs to locate sprinklers whose depth was missing, cut cross-sections through the pipe runs, and checked fitted routes by projecting them onto the photographs (Fig. 6h). With TLS (C4, C5), they sliced the point cloud by height and along routes (Fig. 6f, i), fitted walls and floors with robust plane fits (residuals of 4–9 mm in S1) and pipes with robust circle and cylinder fits, and, in C5, projected the point cloud into the posed panoramas to read classes and connections from the images and rendered the finished model from the same stations for checking (Fig. 6g). With a prior model (C1+P, C2+P), the agents parsed the IFC file, kept its geometry and coordinates, placed the components seen in the images relative to its walls, doors, and columns, and rendered the model from the estimated camera poses to check the alignment (Fig. 6b).

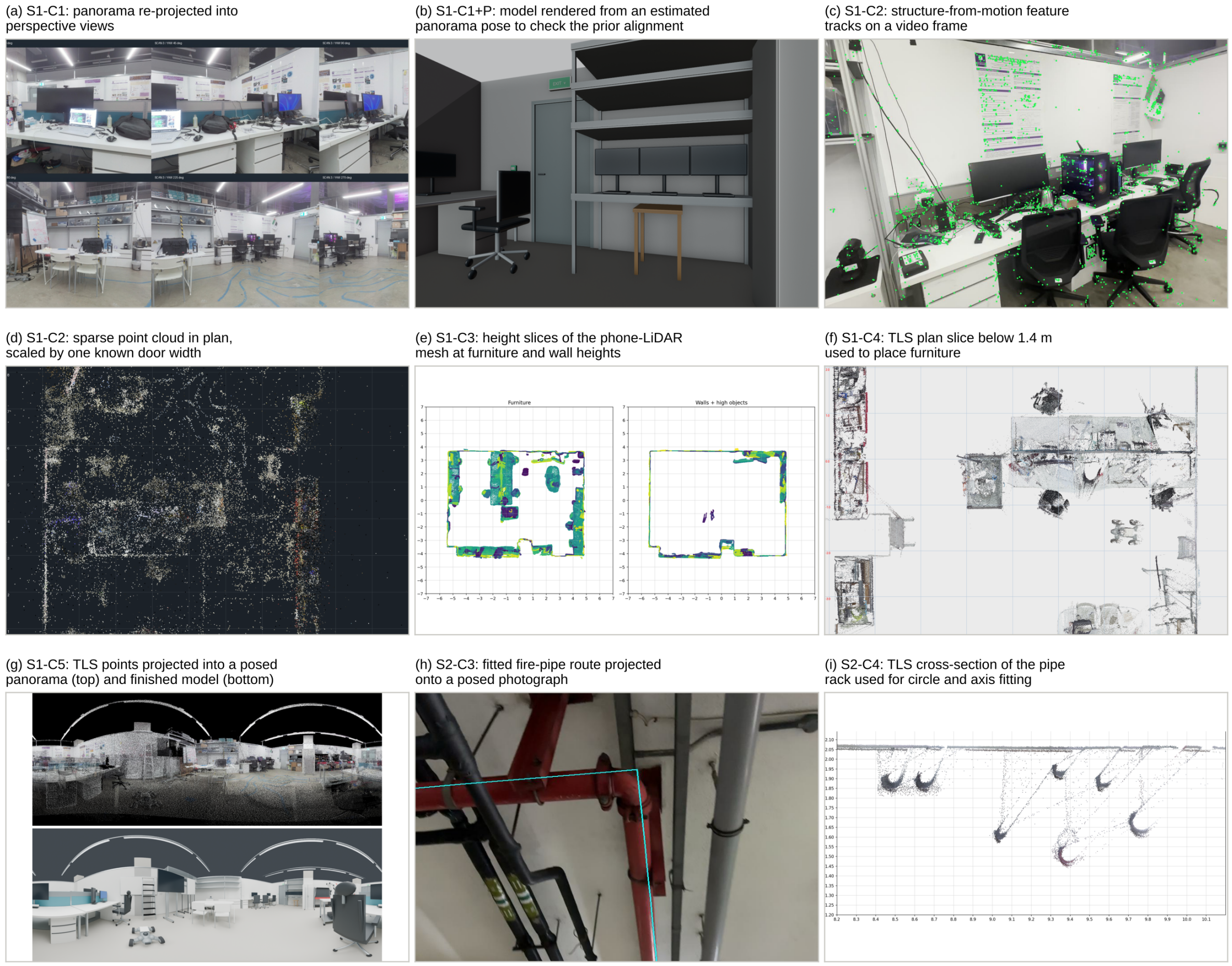

**Fig. 6** Intermediate products generated by the agents, illustrating their strategies: (a) perspective views re-projected from a panorama (S1, C1); (b) model rendered from an estimated panorama pose to check the alignment with the prior (S1, C1+P); (c) structure-from-motion feature tracks on a video frame and (d) the scaled sparse point cloud in plan (S1, C2); (e) height slices of the phone-LiDAR mesh (S1, C3); (f) TLS plan slice below 1.4 m (S1, C4); (g) TLS points projected into a posed panorama (top) and the finished model rendered from the same station (bottom) (S1, C5); (h) fitted fire-pipe route projected onto a posed photograph (S2, C3); (i) TLS cross-section of the pipe rack used for circle and axis fitting (S2, C4).

The summaries also show that the agents were candid about uncertainty. Every scale-free model was declared an estimate whose quantities, positions, and dimensions had not been verified, and the C1 agent in S1 noted that the reference width alone implied a scale uncertainty of ±2.86% before any other error. The agents also documented their counting conventions, such as counting a double door as one door and a continuous shelving run as one shelf, and flagged components whose identity they could not confirm.

## 5.2 Scene S1: General Indoor Scene

Fig. 7 shows the reference model and the models of all configurations in one synchronized view, and Fig. 8 and Table 9 summarize the evaluation. In S1, recognition, as measured by the detection metrics, depended only weakly on the capture modality. From five panoramas or a single video, the agents detected 100 and 98 of the 103 reference components (recall 97.1% and 95.1%), and every configuration except TLS alone reached a recall of at least 92%. Position, by contrast, depended strongly on metric evidence; because detection is counted without a distance limit, many components that count as detected in the scale-free models lie far from their true position. The median position error of the detected components was 61.1 cm for C1 and 25.5 cm for C2, but 8.8 cm for the smartphone LiDAR scan (C3) and 4.8 cm for TLS (C4). Consequently, only 9 of the 102 components evaluated for position (8.8%) lay within 20 cm of their true position in the C1 model and 39 (38.2%) in the C2 model, against 88 (86.3%) for C3 and 76 (74.5%) for C4. The mean errors over the 79 components that every run matched lead to the same ranking (Fig. 8c).

**Table 8** Modeling strategies observed in the agents' code and summaries.

| Config. | Method / core strategy | Source of scale and frame | Main limitations reported by the agent |
|---|---|---|---|
| C1 | Perspective re-projection of panoramas; cross-view identification; ray–floor intersection; two-view triangulation (S2) | Reference door width; assumed level floor and camera height | Automatic pose recovery failed on low texture (S2); layout regularized; overhead components placed at estimated heights |
| C1+P | IFC parsing; components placed relative to IFC walls, doors, and columns; model rendered from estimated panorama poses | Prior IFC frame | Prior furniture and lights inaccurate (S1); no MEP in the prior (S2) |
| C2 | Frame extraction and structure from motion; keyframe inspection; scaling of the sparse cloud by one dimension | Reference door width (S1) or pipe length (S2) | Fragmented reconstruction under a low-texture ceiling (S2); local drift |
| C2+P | IFC parsing; keyframe inspection; structure from motion only for local relations (S2); in S1, placement by visual judgment relative to IFC elements | Prior IFC frame | Conflicts between prior and images; prior furniture adopted where the video was ambiguous (S1) |
| C3 | Mesh height slices and cross-sections; depth back-projection of image points; cylinder fitting; multi-view ray intersection; projection of routes onto photographs | LiDAR scale and ARKit poses | Missing depth on sprinklers; mesh blurs thin pipes |
| C4 | Height slices, elevations, and route profiles of the point cloud; robust plane, circle, and cylinder fitting | Native TLS frame | Small and dark objects hard to identify without images |
| C5 | As C4, plus projection of the point cloud into the posed panoramas and rendering of the model from the stations | Native TLS frame and panorama poses | Panorama orientation convention had to be recovered |

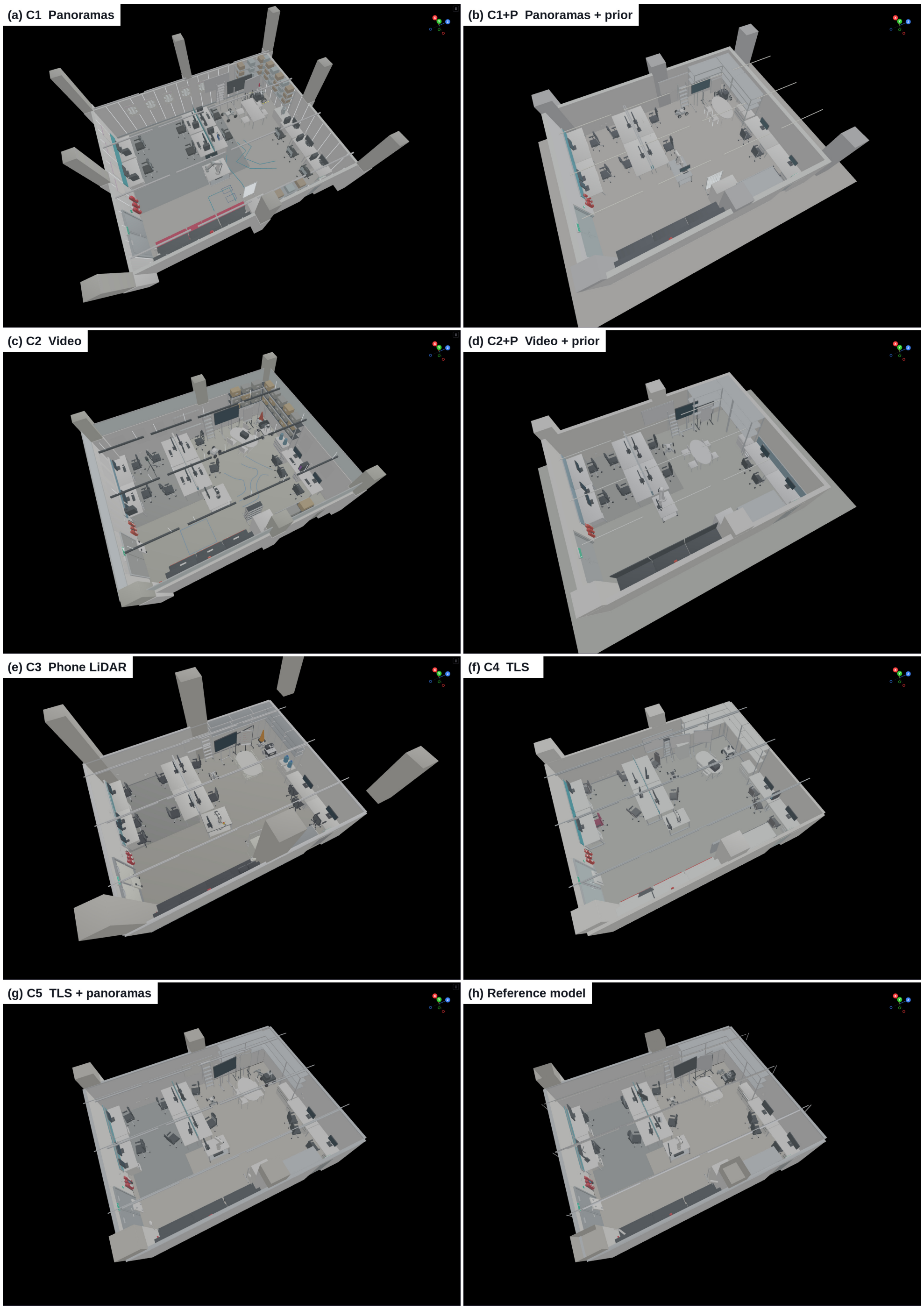

**Fig. 7** Synchronized view of scene S1 in a common frame: (a)–(g) agent-generated models of C1, C1+P, C2, C2+P, C3, C4, and C5; (h) reference model.

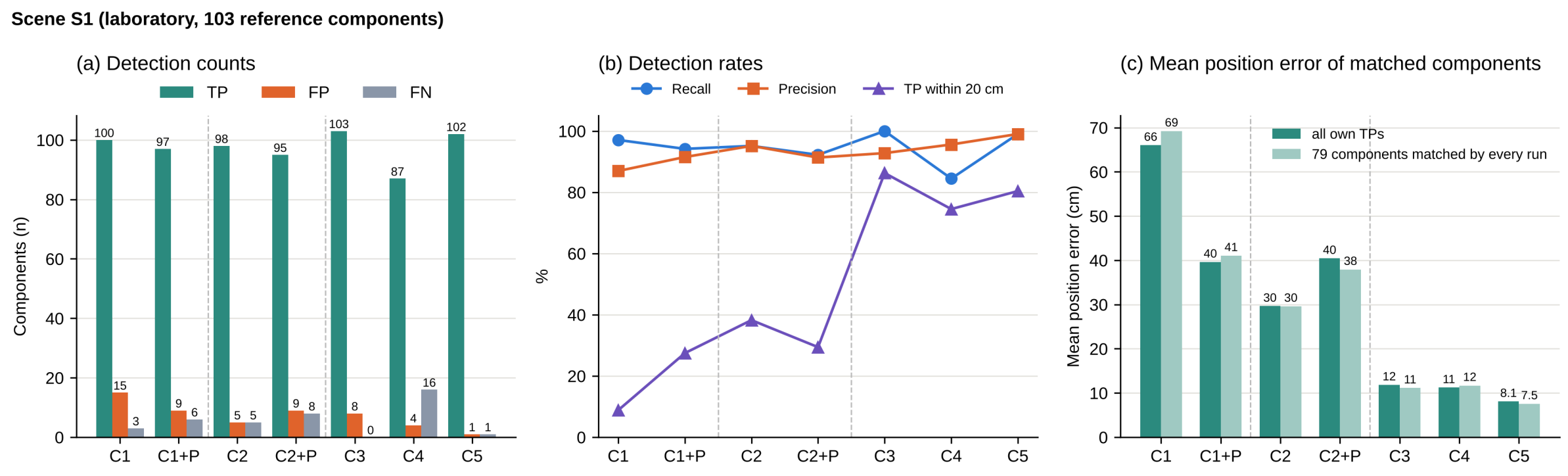

**Fig. 8** Results in scene S1: (a) true positives, false positives, and false negatives; (b) recall, precision, and share of reference components detected within 20 cm; (c) mean position error of the true positives, over each run's own matches and over the 79 components matched by every run (floor excluded from position statistics). Dashed lines separate the panorama-based, video-based, and metric configurations.

**Table 9** Detection and position accuracy in scene S1 (103 reference components). Shading: darker is better within each column.

| Config. | Modeled (n) | TP | FP | FN | Recall (%) | Precision (%) | F1 | Median error (cm) | Mean error (cm) | P90 error (cm) | TP within 20 cm, n (% of 102) |
|---|---|---|---|---|---|---|---|---|---|---|---|
| C1 | 115 | 100 | 15 | 3 | 97.1 | 87.0 | 0.917 | 61.1 | 66.0 | 117.1 | 9 (8.8) |
| C1+P | 106 | 97 | 9 | 6 | 94.2 | 91.5 | 0.928 | 35.7 | 39.6 | 80.4 | 28 (27.5) |
| C2 | 103 | 98 | 5 | 5 | 95.1 | 95.1 | 0.951 | 25.5 | 29.7 | 60.8 | 39 (38.2) |
| C2+P | 104 | 95 | 9 | 8 | 92.2 | 91.3 | 0.918 | 31.4 | 40.5 | 79.0 | 30 (29.4) |
| C3 | 111 | 103 | 8 | 0 | 100.0 | 92.8 | 0.963 | 8.8 | 11.9 | 25.0 | 88 (86.3) |
| C4 | 91 | 87 | 4 | 16 | 84.5 | 95.6 | 0.897 | 4.8 | 11.2 | 20.8 | 76 (74.5) |
| C5 | 103 | 102 | 1 | 1 | 99.0 | 99.0 | 0.990 | 0.7 | 8.1 | 26.1 | 82 (80.4) |

Position error: horizontal distance between bounding-box centers after rigid registration (3D distance for wall-mounted devices). Position statistics exclude the floor slab (102 components).

Two contrasts isolate the roles of geometry and imagery. TLS without images (C4) missed 16 components, more than any other configuration: 7 of the 16 monitors, 4 of the 6 wall sockets, 2 cabinets, a quadruped robot, a fire-safety device, and a table (Table 10). Dark screens return few laser points, and small faceplates are hard to distinguish from the wall in a point cloud. Adding the posed panoramas of the same stations (C5) recovered almost all of them, and recall rose from 84.5% to 99.0%; this gain is partly inflated by the derivation of the reference from C5, but the missed classes are those that are difficult to resolve from the point cloud alone, whose colors are sparse and of low resolution compared with photographs. Conversely, the same panoramas without poses or point cloud (C1) supported recognition but not position. The smartphone LiDAR scan (C3) combined both kinds of evidence in one capture: it detected all 103 components, with eight false positives, and placed 86% of them within 20 cm.

**Table 10** Detection per class in scene S1 (TP / reference components; false positives in parentheses where present). Shading: darker = higher share detected.

| Class | C1 | C1+P | C2 | C2+P | C3 | C4 | C5 |
|---|---|---|---|---|---|---|---|
| Walls | 4/4 | 4/4 | 4/4 | 4/4 | 4/4 | 4/4 | 4/4 |
| Columns | 4/4 (4) | 4/4 | 4/4 (2) | 4/4 | 4/4 | 4/4 | 4/4 |
| Floor | 1/1 | 1/1 | 1/1 | 1/1 | 1/1 | 1/1 | 1/1 |
| Doors | 2/2 | 2/2 | 2/2 | 2/2 | 2/2 | 2/2 | 2/2 |
| Tables | 14/14 (2) | 14/14 | 14/14 | 14/14 | 14/14 (1) | 13/14 | 14/14 |
| Chairs | 13/13 | 11/13 | 13/13 | 13/13 | 13/13 (3) | 13/13 | 13/13 |
| Shelves | 2/2 | 2/2 | 2/2 | 2/2 | 2/2 (1) | 2/2 | 2/2 |
| Cabinets | 11/11 | 11/11 | 10/11 | 10/11 (2) | 11/11 | 9/11 (1) | 11/11 |
| Whiteboards | 3/3 | 3/3 | 3/3 | 3/3 | 3/3 | 3/3 | 3/3 |
| Ladders | 1/1 | 1/1 | 1/1 | 1/1 | 1/1 | 1/1 | 1/1 |
| Trolley | 0/1 | 0/1 | 1/1 | 0/1 | 1/1 | 1/1 | 1/1 |
| Monitors | 14/16 (1) | 15/16 | 15/16 | 15/16 | 16/16 (1) | 9/16 | 15/16 |
| Robot arms | 1/1 | 1/1 | 1/1 | 1/1 | 1/1 | 1/1 | 1/1 |
| Mobile robots | 1/1 | 1/1 | 1/1 | 0/1 | 1/1 | 1/1 (1) | 1/1 |
| Quadruped robots | 2/2 | 1/2 | 2/2 | 1/2 | 2/2 | 1/2 | 2/2 |
| Lighting | 12/12 (4) | 12/12 (9) | 12/12 | 12/12 (6) | 12/12 | 12/12 | 12/12 |
| Wall sockets | 6/6 (4) | 5/6 | 5/6 (1) | 5/6 (1) | 6/6 (2) | 2/6 (2) | 6/6 (1) |
| Fire extinguishers | 3/3 | 3/3 | 3/3 | 3/3 | 3/3 | 3/3 | 3/3 |
| Other fire-safety equipment | 4/4 | 4/4 | 2/4 (2) | 2/4 | 4/4 | 3/4 | 4/4 |
| Exit signs | 2/2 | 2/2 | 2/2 | 2/2 | 2/2 | 2/2 | 2/2 |

Fig. 9 shows where these differences arise. In the plan views of the matched components, each true positive is joined to its reference counterpart by a line whose length is the position error. In the scale-free models, the lines grow toward the far end of the room and fan out in the furniture clusters, which indicates a global scale and layout error rather than random misplacement; in the metric models, they shrink to points.

The architecture shows the same pattern at the scale of the room (Fig. 10 and Table 11). Both scale-free models underestimated the room despite the reference width: the floor envelope was 9.60 m × 7.70 m (C1) and 9.68 m × 7.60 m (C2) against 10.15 m × 7.92 m, that is, 4.6–5.4% too short and 2.8–4.0% too narrow. The smartphone LiDAR model was within 1% (10.25 m × 7.95 m). The C1 agent also modeled eight columns instead of four, which accounts for four of its fifteen false positives. With the prior, the walls and columns came from the IFC model, which is approximately correct for the architecture, and their center offsets fell from 18–20 cm to 6 cm and from 43–60 cm to 16–17 cm.

The other categories follow the same split, with differences that reflect how each kind of component is observed (Fig. 11). Furniture was placed within about 20 cm on average by the metric configurations, with two exceptions (whiteboards in C4, 61 cm, and cabinets in C5, 21 cm), but mostly 20–70 cm away by the scale-free ones; chairs were among the least accurate classes in every metric model (13–19 cm), because they are movable, rotated, and partly occluded. For displays and robots, C1 erred by 39–77 cm, whereas C3 placed every class within about 14 cm; TLS alone placed the robots within 6 cm but found only 9 of the 16 monitors. Among building services, ceiling lights were the most sensitive to scale: their mean error ranged from 107 cm for C1 to 11 cm for C2, whose structure from motion constrained the ceiling, and 1–4 cm for the metric models. Wall sockets were the hardest class for every input except C5: small faceplates near the floor were placed with mean errors of 33 cm (C3) and 49 cm (C4), and TLS alone found only 2 of 6. Fire extinguishers, other fire-safety devices, and exit signs were found by almost every configuration, with errors of 13–32 cm for the scale-free inputs and at most 16 cm for the metric ones.

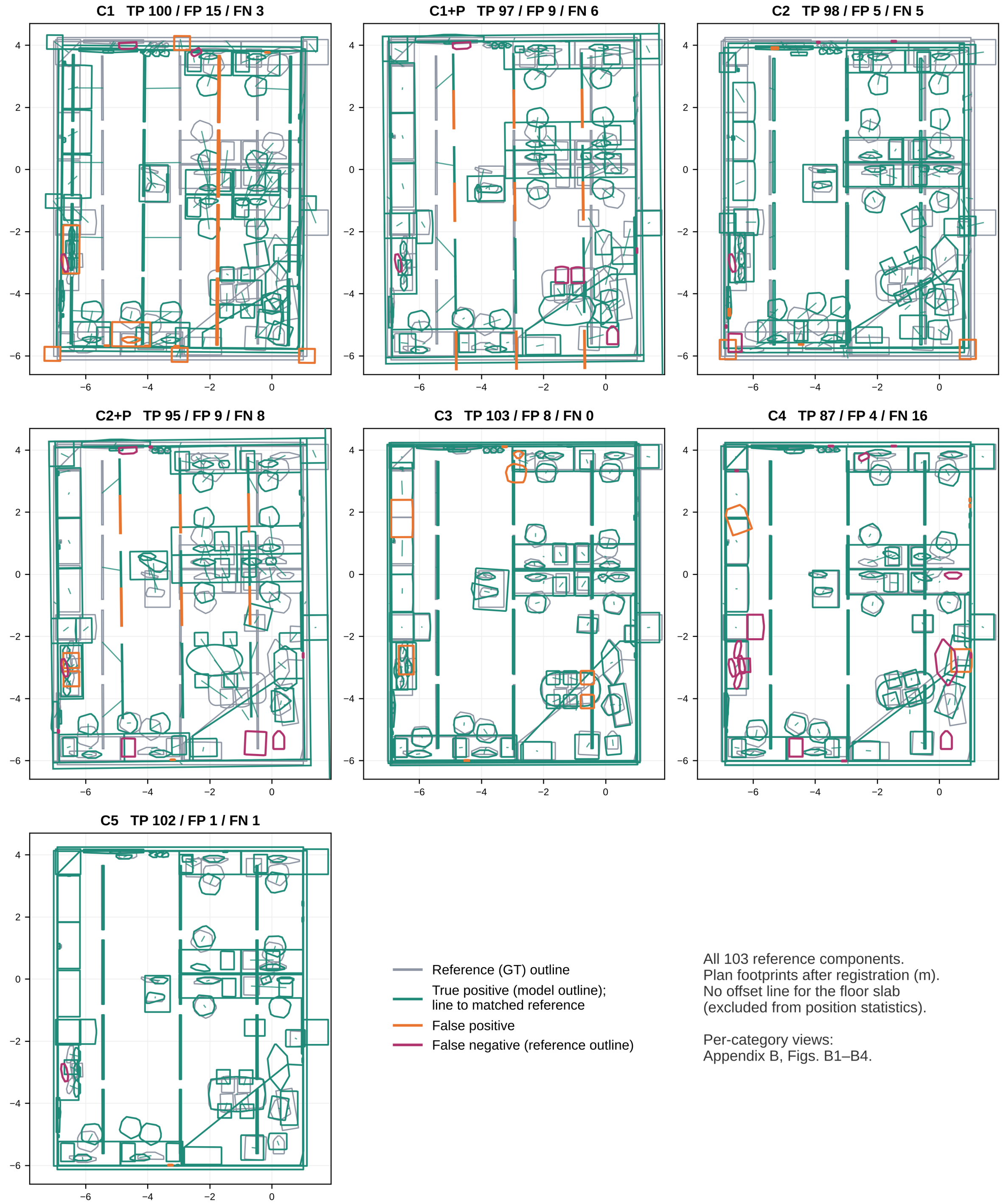

**Fig. 9** Plan footprints of the reference model and the agent-generated models in scene S1 after registration: reference components (gray), true positives (teal), false positives (orange), and false negatives (purple). Lines join the center of each true positive to that of its matched reference component. Views per component category are given in Appendix B (Figs. B1–B4).

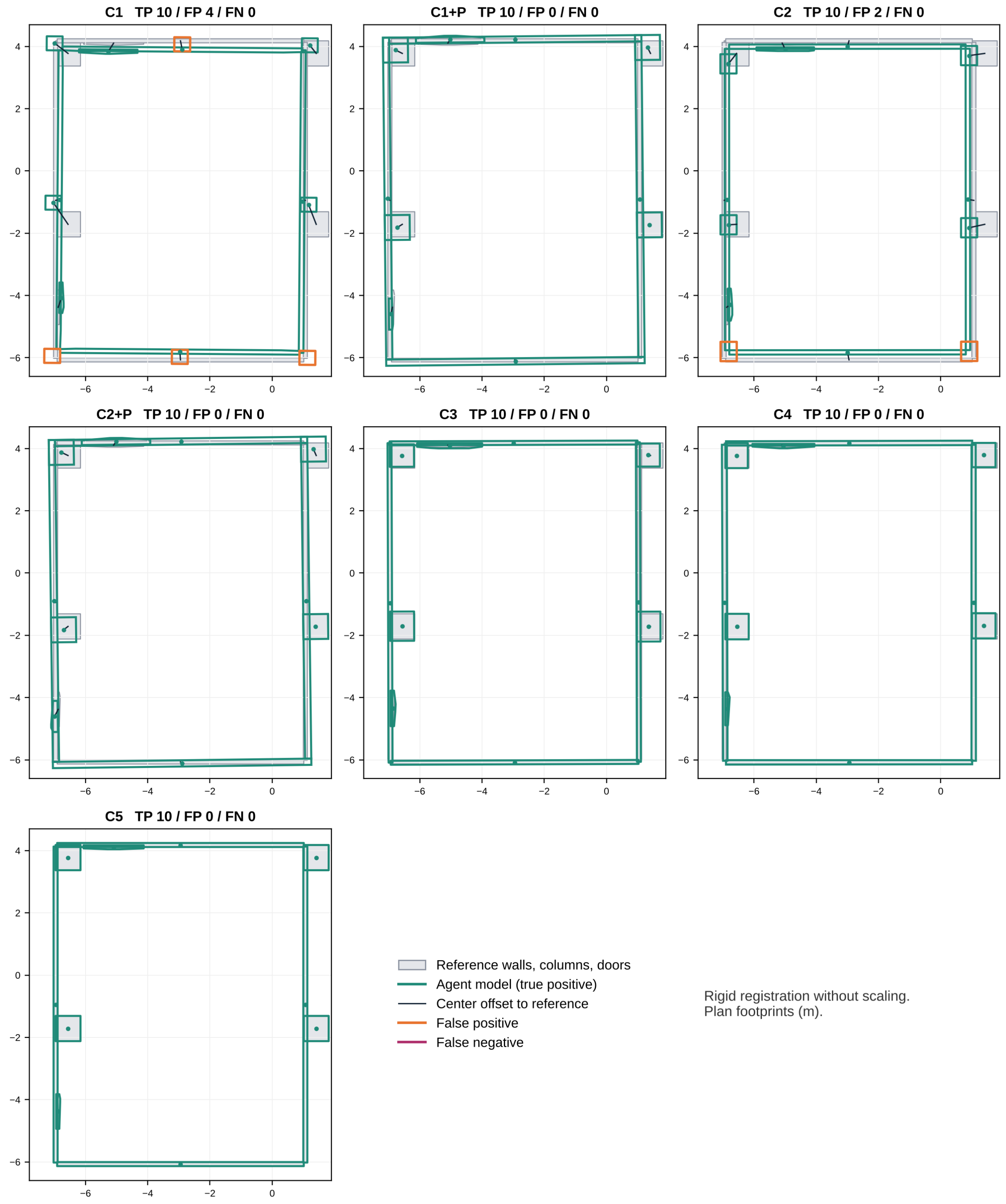

**Fig. 10** Walls, columns, and doors of scene S1 in plan after registration: reference components (gray fill), matched agent-generated components (teal), false positives (orange), and false negatives (purple). Lines join the center of each matched component to that of its reference component.

**Table 11** Architectural components in scene S1. Shading: darker is better within each column.

| Config. | Floor envelope L × W (m) | Wall center offset (cm) | Column center offset (cm) | Door center offset (cm) | Wall dimension error (cm) | Column dimension error (cm) |
|---|---|---|---|---|---|---|
| Reference | 10.15 × 7.92 | — | — | — | — | — |
| C1 | 9.60 × 7.70 | 20 | 60 | 32 | 17 | 68 |
| C1+P | 14.74 × 10.11 † | 6 | 17 | 18 | 10 | 4 |
| C2 | 9.68 × 7.60 | 18 | 43 | 18 | 35 | 35 |
| C2+P | 14.74 × 10.11 † | 6 | 16 | 20 | 10 | 4 |
| C3 | 10.25 × 7.95 | 3 | 4 | 5 | 3 | 63 |
| C4 | 10.14 × 7.91 | 1 | 2 | 4 | 3 | 2 |
| C5 | 10.15 × 7.92 | 0 | 0 | 0 | 0 | 1 |

† In C1+P and C2+P, the agents adopted the floor slab of the prior IFC model unchanged. This slab was drawn roughly from the floor plan and extends beyond the walls of the room, so its envelope (14.74 m × 10.11 m) is not a reconstruction of the floor and is not comparable with the other rows.

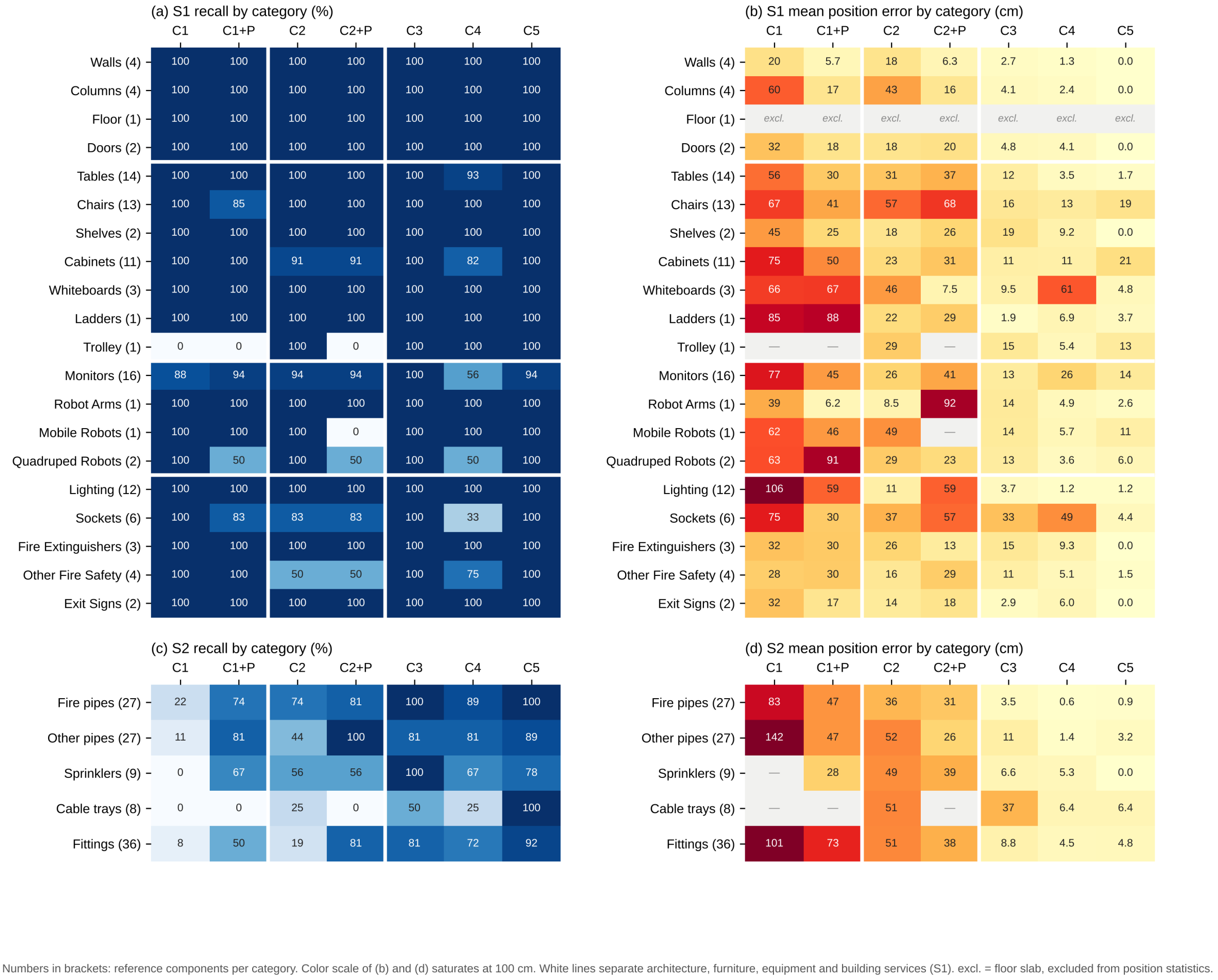

**Fig. 11** Recall and mean position error per category: (a), (b) scene S1; (c), (d) scene S2. Gray cells: no true positive. The floor slab is counted in recall but excluded from the position error (excl.).

## 5.3 Scene S2: Multi-System MEP Scene

The MEP scene separated the configurations far more sharply (Figs. 12–14 and Table 12). The scale-free inputs failed from scratch: the C1 model matched only 12 of the 107 reference components (recall 11.2%, precision 18.5%), and the C2 model 46 (recall 43.0%, precision 63.0%). Their matched components were displaced by a median of 113 cm and 46 cm, the C1 pipes deviated in direction by a median of 35°, and only 7.2% and 14.5% of the reference connections were recovered. With a single reference dimension and no metric depth, the agents could not place an overhead network: ray–floor intersections, which worked for furniture standing on the floor in S1, do not constrain components under a ceiling, and the low texture of walls and ceiling defeated automatic pose recovery.

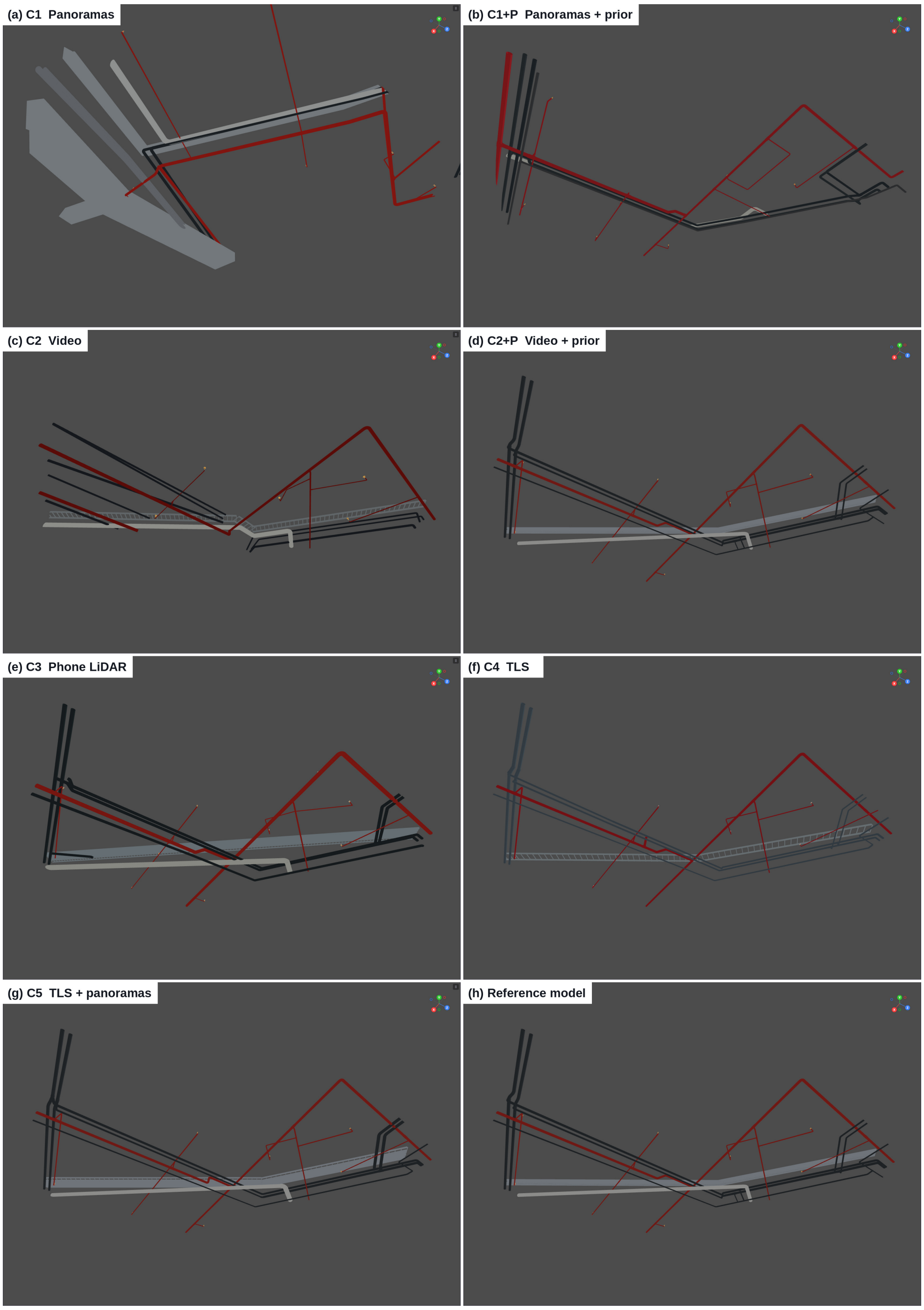

**Fig. 12** Synchronized view of scene S2 in a common frame: (a)–(g) agent-generated MEP models of C1, C1+P, C2, C2+P, C3, C4, and C5; (h) reference model.

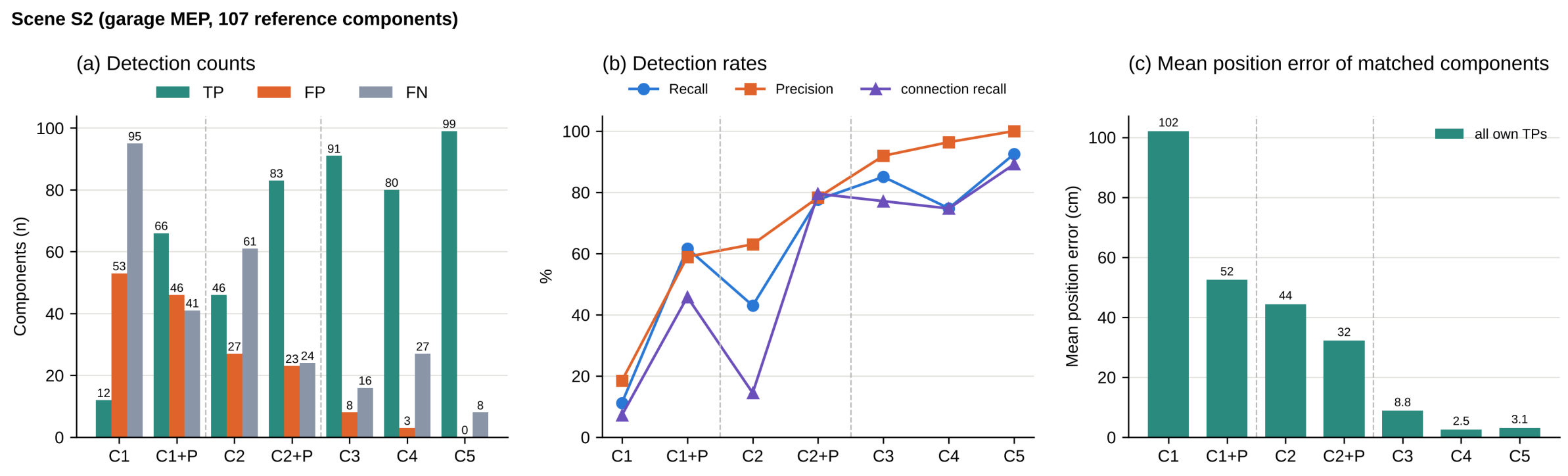

**Fig. 13** Results in scene S2: (a) true positives, false positives, and false negatives; (b) recall, precision, and connection recall; (c) mean position error of the true positives. Dashed lines separate the panorama-based, video-based, and metric configurations.

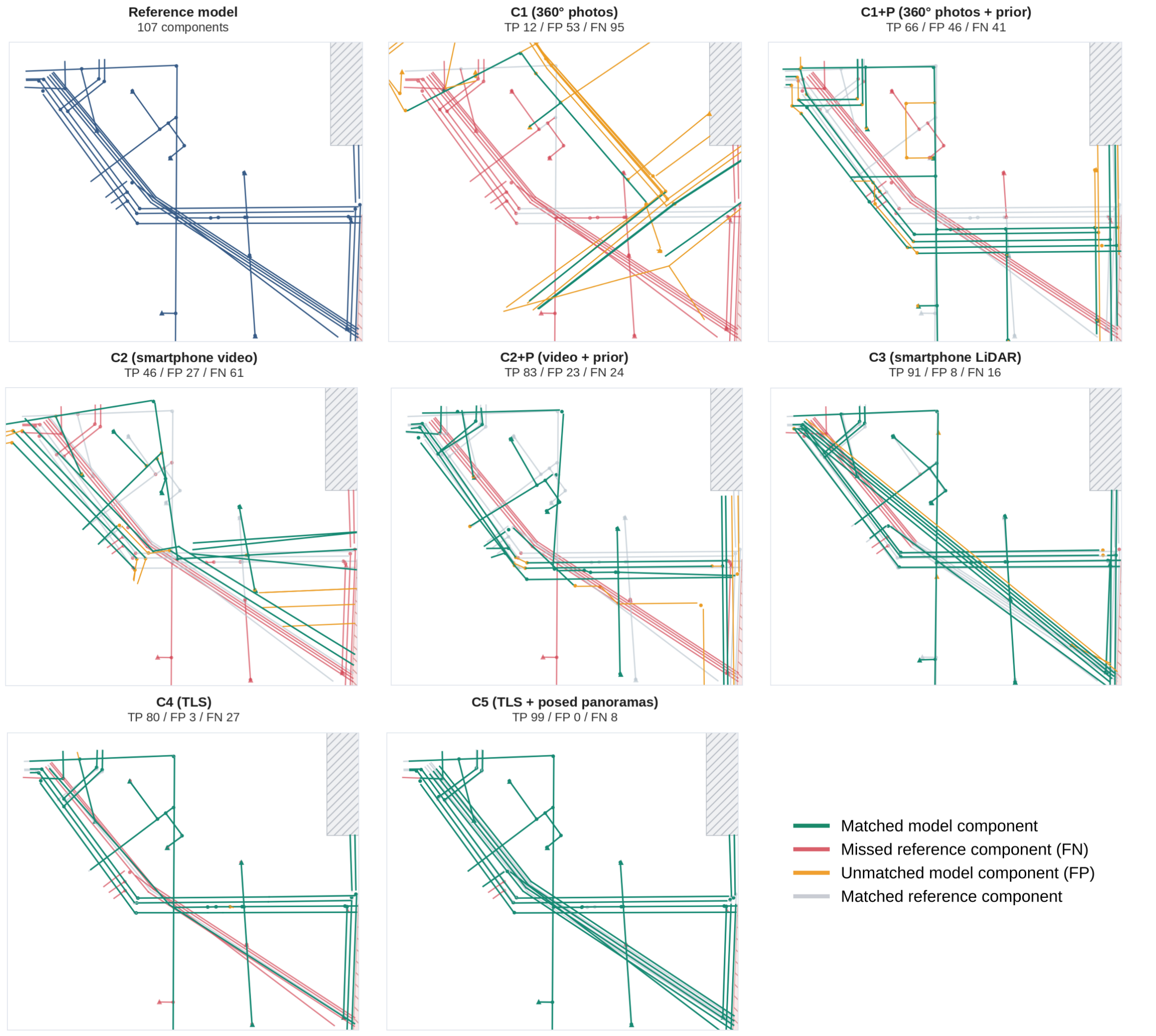

**Fig. 14** Plan views of the evaluated MEP components in scene S2 for the reference model and the seven configurations, with matched, missed, and unmatched components; hatched areas lie outside the evaluated region. Detailed maps per configuration are given in Appendix B (Figs. B5–B11).

**Table 12** Detection, geometry, and topology in scene S2 (107 reference components). Shading: darker is better within each column.

| Config. | TP | FP | FN | Recall (%) | Precision (%) | F1 | Median position error (cm) | Median direction error (°) | Median diameter error (%) | Length MAE (cm) | Connection recall (%) |
|---|---|---|---|---|---|---|---|---|---|---|---|
| C1 | 12 | 53 | 95 | 11.2 | 18.5 | 0.14 | 112.8 | 35.2 | 50.8 | 64 | 7.2 |
| C1+P | 66 | 46 | 41 | 61.7 | 58.9 | 0.60 | 43.5 | 1.6 | 12.0 | 66 | 45.8 |
| C2 | 46 | 27 | 61 | 43.0 | 63.0 | 0.51 | 46.1 | 7.0 | 23.7 | 57 | 14.5 |
| C2+P | 83 | 23 | 24 | 77.6 | 78.3 | 0.78 | 30.7 | 2.7 | 23.1 | 37 | 79.5 |
| C3 | 91 | 8 | 16 | 85.0 | 91.9 | 0.88 | 4.9 | 1.0 | 23.1 | 16 | 77.1 |
| C4 | 80 | 3 | 27 | 74.8 | 96.4 | 0.84 | 1.0 | 0.3 | 7.2 | 10 | 74.7 |
| C5 | 99 | 0 | 8 | 92.5 | 100.0 | 0.96 | 0.0 | 0.0 | 0.0 | 13 | 89.2 |

Geometric errors are computed over true positives only; position error is the mean distance between matched centerlines.

The near-zero geometric errors reported for C5 should be interpreted in light of the reference-model construction procedure (Section 4.3), as the reference models were refined from the C5 outputs rather than developed independently.

The metric configurations recovered most of the network. The smartphone LiDAR scan (C3) matched 91 components (recall 85.0%, precision 91.9%) with a median position error of 4.9 cm, a median direction error of 1.0°, and 77.1% of the connections. TLS alone (C4) was the most precise geometrically, with a median position error of 1.0 cm, a direction error of 0.3°, and a diameter error of 7.2%, but it matched fewer components (80, recall 74.8%) than C3: it found only 2 of the 8 cable-tray sections and 6 of the 9 sprinklers (Table 13), which in the point cloud alone are hard to tell apart from the trunking, hangers, and parallel pipes around them. Adding the posed panoramas (C5) again recovered them, as in S1. C3 found all 27 fire-protection pipes and all nine sprinklers, which suggests that the close-range photographs of the phone scan helped with identification, although its outer diameters, taken from a mesh that blurs thin pipes, were less accurate (23% median error). By category (Fig. 11c, d), fire-protection pipes were placed within 4 cm on average by every metric configuration, whereas cable trays remained the weakest category for all inputs except C5, and the fitting groups, whose position depends on the correct junction of two pipes, were placed within 9 cm by the metric configurations but 38–101 cm by the scale-free ones.

**Table 13** Detection per category in scene S2 (TP / reference components; false positives in parentheses where present). Shading: darker = higher share detected.

| Category | C1 | C1+P | C2 | C2+P | C3 | C4 | C5 |
|---|---|---|---|---|---|---|---|
| Fire-protection pipe sections | 6/27 (12) | 20/27 (12) | 20/27 (1) | 22/27 (2) | 27/27 | 24/27 (2) | 27/27 |
| Other service pipe sections | 3/27 (11) | 22/27 (7) | 12/27 (10) | 27/27 (8) | 22/27 (1) | 22/27 | 24/27 |
| Sprinklers | 0/9 (5) | 6/9 (1) | 5/9 | 5/9 | 9/9 (2) | 6/9 | 7/9 |
| Cable-tray sections | 0/8 (6) | 0/8 | 2/8 | 0/8 | 4/8 (1) | 2/8 | 8/8 |
| Fitting groups | 3/36 (16) | 18/36 (26) | 7/36 (16) | 29/36 (13) | 29/36 (4) | 26/36 (1) | 33/36 |

Within the range of evaluation tolerances that was examined, the S2 counts did not change, because the correspondences are established by the network context rather than by the tolerances. Appendix B shows, for each configuration, where the matched, missed, and false components lie, how the position offset is distributed along the network, and which local connections were recovered (Figs. B5–B11). These maps were produced by the evaluation code from the same match tables as Tables 12 and 13, so they agree with the reported values.

## 5.4 Effect of the Prior Model

The prior model had a large effect in the MEP scene and a mixed one in the laboratory (Table 14). In S2, the prior raised the F1 score from 0.14 to 0.60 for the panoramas and from 0.51 to 0.78 for the video, cut the median direction error from 35° to 1.6° and from 7.0° to 2.7°, and raised the connection recall from 7% to 46% and from 14% to 80%. The prior shares the frame of the scan and fixes the walls, door openings, and ceiling to which the services run parallel, so the agents no longer had to infer the frame from a single dimension. The prior did not, however, tell the agents where each pipe lies within that frame: the median position errors remained 44 cm and 31 cm, an order of magnitude above those of the metric configurations.

In S1, the effect depended on which part of the prior was accurate. The S1 prior was drawn roughly by hand from the floor plan: its architecture is approximately correct, but its furniture and light fittings were placed without measurement. Where the prior was accurate, it helped both scale-free inputs: the wall and column offsets fell to 6 cm and 16–17 cm (Table 11). Where it was not, it hurt. For C1, whose own positions were poor, the prior still improved the overall result: the median position error fell from 61.1 cm to 35.7 cm and the share of components within 20 cm rose from 8.8% to 27.5%. For C2, whose structure from motion had already placed the contents reasonably, the inaccurate furniture and lights of the prior made the result worse: the median error rose from 25.5 cm to 31.4 cm, the share within 20 cm fell from 38.2% to 29.4%, and precision fell from 95.1% to 91.3%. The prior lights produced six to nine false-positive or misplaced lights (mean light error 59 cm with the prior against 11 cm for C2 alone), and the C2+P agent placed much of the furniture relative to the IFC elements instead of reconstructing it from the video. An accurate prior would therefore be expected to improve the scale-free configurations substantially, in the contents as well as in the fabric; a rough one helps only where it is right and can override better evidence where it is wrong.

**Table 14** Paired comparison with and without the prior model.

| Scene | Pair | Δ recall (pp) | Δ precision (pp) | F1 | Median position error (cm) | S1: TP within 20 cm (%); S2: connection recall (%) |
|---|---|---|---|---|---|---|
| S1 | C1 → C1+P | -2.9 | +4.6 | 0.917 → 0.928 | 61.1 → 35.7 | 8.8 → 27.5 |
| S1 | C2 → C2+P | -2.9 | -3.8 | 0.951 → 0.918 | 25.5 → 31.4 | 38.2 → 29.4 |
| S2 | C1 → C1+P | +50.5 | +40.5 | 0.140 → 0.603 | 112.8 → 43.5 | 7.2 → 45.8 |
| S2 | C2 → C2+P | +34.6 | +15.3 | 0.511 → 0.779 | 46.1 → 30.7 | 14.5 → 79.5 |

## 5.5 Efficiency and Cost

Table 15 lists the effort per run. The agent effort varied far less than the capture effort: a run took 17–53 min, 30–85 tool calls, 3.0–10.7 million tokens, and USD 6–16 of API cost, and all fourteen runs together cost about USD 147 and 7.8 h of unattended agent time. On site, by contrast, the configurations differ by more than an order of magnitude, from 2–3 min for a video (measured from the recordings) or a smartphone LiDAR scan to 20–30 min for a TLS survey (approximate on-site times). For the 360° panoramas, about 1 min is an estimate for a consumer 360° camera; in this paper, the panoramas were recorded by the TLS during its survey. Because the agent cost was of the same order for every input in these runs, the trade-off between configurations was governed mainly by capture time and accuracy.

Read in this way, the smartphone LiDAR scan (C3) offered the most favorable trade-off in these experiments: for 2–3 min on site and about USD 13 per run, it gave the highest recall in S1 (100%) and 85% recall in S2, with median position errors of 8.8 cm and 4.9 cm. TLS-based configurations (C4 and C5) achieved the lowest geometric errors, although C5's results are influenced by the reference-model construction procedure. TLS capture took approximately ten times longer than smartphone capture, but its agent runs were not more expensive than the others, and C4 was the cheapest metric run in both scenes (USD 7.7 and 9.8). The scale-free inputs were the fastest to capture but required a human-supplied reference dimension and, in S2, did not deliver a usable network. The prior shortened the S1 runs to 17–21 min, the shortest of all, because the agents reused the IFC geometry; in S2, the C2+P run was the longest and most expensive (53 min, USD 16.4), as the agent reconciled the video with the IFC frame route by route.

**Table 15** Efficiency and cost per configuration (S1 / S2).

| Config. | On-site capture (min) | Agent runtime (min) | Tool calls (n) | Tokens (k) | API cost (USD) | Operator inputs (n) |
|---|---|---|---|---|---|---|
| C1 | ≈ 1 | 26 / 40 | 38 / 62 | 4,440 / 7,952 | 7.97 / 13.16 | 1 |
| C1+P | ≈ 1 | 21 / 26 | 34 / 33 | 3,721 / 3,831 | 7.13 / 6.62 | 0 |
| C2 | 2–3 | 35 / 35 | 57 / 60 | 7,437 / 6,676 | 12.15 / 10.68 | 1 |
| C2+P | 2–3 | 17 / 53 | 30 / 85 | 2,983 / 10,713 | 6.12 / 16.39 | 0 |
| C3 | 2–3 | 38 / 34 | 59 / 58 | 8,997 / 7,439 | 13.80 / 12.71 | 0 |
| C4 | 20–30 | 35 / 28 | 37 / 46 | 4,206 / 5,636 | 7.68 / 9.84 | 0 |
| C5 | 20–30 | 31 / 48 | 40 / 58 | 5,100 / 8,223 | 9.19 / 13.53 | 0 |

Values are given as S1 / S2. On-site capture times as in Table 3: C1 and C1+P estimated, C2 and C2+P measured from the video durations, C3–C5 approximate times recorded on site. Operator inputs refer to the reference dimensions provided in response to initial agent requests for C1 and C2. No additional operator inputs or manual corrections were recorded during model generation. Each value comes from a single run.

# 6. DISCUSSION

## 6.1 How AI-driven Modeling Reshapes Scan-to-BIM

The results point to five shifts in how Scan-to-BIM may be organized when an agent performs the modeling. They are observations from two scenes and one run per configuration and should be read as hypotheses to be confirmed by broader validation rather than as general conclusions.

**From modeling labor to evidence sufficiency.** Conventional pipelines are limited by what they can recognize; the agents in this study detected 92–100% of the laboratory components from every input that contained images, including five panoramas and a two-minute video, and they did so for open-vocabulary classes such as robots, whiteboards, and fire-safety devices without training data. Where the inputs differed was in locating the components: the share of components placed within 20 cm of their true position ranged from 9% to 86%, depending on whether and how the input carried metric information. In other words, once an agent does the modeling, a capture that shows every component is not enough; its value lies in the metric evidence it provides about where the components are. Capture should therefore be planned around that metric evidence for the components that matter, rather than around visual coverage alone.

**Images tell what, metric data tell where.** The paired configurations show that the two kinds of evidence are complementary. The panoramas without metric data (C1) located components an order of magnitude worse than any metric configuration (median 61 cm against at most 8.8 cm in S1), whereas the point cloud without images (C4) located components precisely but missed those that are hard to identify without photographs, such as monitors, sockets, sprinklers, and cable trays; adding the posed panoramas (C5) raised recall by 15 and 18 percentage points in the two scenes. Because the reference was derived from C5, this gain is likely overstated, but the classes that C4 missed are those that are ambiguous in a point cloud whose colors are sparse and of low resolution. In both scenes, the smartphone LiDAR scan, which records both kinds of evidence, offered the most favorable balance among the low-cost options tested

(Section 6.2).

**Scene type changes what low-cost capture can deliver.** In the furnished laboratory, scale-free imagery produced inventory-grade models: almost every component was found, with positions within a few decimeters. Under the garage ceiling, the same inputs found only 11–43% of the MEP components and few of their connections. Overhead linear networks are elevated, repetitive, and low in texture, and their identity depends on connections, so they need metric depth more than any other component type.

**From scanning everything to anchoring on priors, with care.** An existing model supplied the frame and the fabric and transformed the MEP results, which supports Scan-to-BIM as a model-updating task that merges with Scan-vs-BIM [3,9,144]. The laboratory shows that the benefit depends on the accuracy of each part of the prior: a prior drawn roughly from the floor plan improved the architecture, but its inaccurate furniture and lights introduced false positives and displaced the contents, and the agent leaned on the prior instead of exploiting its own evidence. An accurate prior would be expected to help much more. Agents should therefore be told how reliable each part of a prior is and be instructed to treat it as a hypothesis to be verified rather than as truth.

**From accuracy claims to verification.** Every agent documented its method, its counting conventions, and its open issues, and none claimed survey accuracy for an estimated model. Such self-reports are valuable for review, but they measure consistency with the agent's own assumptions, not accuracy. Verification against measured site data, rather than relying solely on the agents' internal checks, remains essential for assessing the reliability of agent-generated BIM models. Furthermore, the code-generated models and associated scripts support reproducibility and traceability of the modeling process.

## 6.2 Smartphone LiDAR as a Practical Middle Ground, and Where TLS Remains Necessary

Among the configurations tested, the smartphone LiDAR scan (C3) offered the most favorable overall balance in both scenes. It is inexpensive, because it runs on a consumer phone that costs roughly an order of magnitude less than a terrestrial laser scanner and needs no survey expertise. It is fast, taking 2–3 min on site in each scene against 20–30 min for TLS. And its accuracy was not low: in S1, it detected all 103 components and placed 86% of them within 20 cm (median 8.8 cm), more than TLS alone (75%), and in S2 it recovered 85% of the MEP components and 77% of their connections with a median position error of 4.9 cm. Its strength comes from recording both kinds of evidence at once: a metric depth map and mesh that fix where things are, and close-range posed photographs that tell the agent what they are, which is the combination that the TLS-only configuration lacked.

These results come from two testing areas of about 10 m × 8 m and 10.5 m × 12.5 m, scanned in a few minutes each. The phone sensor has a short range of a few meters, its tracking drifts over long trajectories, and its mesh smooths thin and distant objects; the 23% median diameter error for pipes in S2 is an early sign of this. Whether the same accuracy holds for larger floors, tall halls, long corridors, or multi-room capture is not established by this study and must be tested before smartphone LiDAR can replace TLS in such settings.

TLS remains preferable for applications requiring high geometric accuracy. The TLS-only configuration (C4) achieved a median position error of 1.0 cm, a direction error of 0.3°, and a diameter error of 7% for pipes in S2, demonstrating the value of survey-grade metric information. Applications with tight tolerances, such as as-built verification and acceptance, clash detection and prefabrication of MEP retrofits, deformation and condition assessment, and documentation that must stand as a record, need this survey-grade accuracy. TLS also covers large, tall, or poorly lit spaces from few stations with uniform accuracy, and it provided the verified reference in this study. The practical division of labor is thus smartphone LiDAR for fast, low-cost modeling of rooms and their contents, and TLS, preferably with posed panoramas, where accuracy, scale, or accountability are critical.

A potential alternative to TLS is advanced mobile laser scanning (MLS). Current high-end mobile and handheld scanners report accuracies of less than 1 cm while capturing much faster than static TLS, and they record not only the final point cloud but also images, depth, and the trajectory with sensor poses, that is, the same combination of metric geometry and posed imagery that made C3 and C5 effective. Such systems are better suited to large scenes, where static TLS becomes slow and smartphone LiDAR loses accuracy, and they are a natural candidate for the next round of experiments.

## 6.3 Failure Modes

Table 16 summarizes the failure modes observed in the fourteen runs (RQ4).

**Table 16** Failure modes of agent-driven Scan-to-BIM observed in this study.

| Failure mode | Description | Evidence | Most affected | Mitigation |
|---|---|---|---|---|
| Scale and layout ambiguity | One reference dimension fixes scale locally but not the layout | S1 envelope 9.60 × 7.70 m (C1) vs. 10.15 × 7.92 m; median error 61 cm (C1), 25 cm (C2) | C1, C2 | Metric capture (C3–C5) or prior model; several reference dimensions |
| Elevated components without depth | Ray–floor intersection cannot place components under the ceiling | S2 recall 11% (C1), 43% (C2); median direction error 35° (C1) | C1, C2 in MEP scenes | Metric depth; prior with ceiling height |
| Pose-recovery failure | Low texture and repetition break structure from motion | S2: automatic pose recovery failed (C1); fragmented reconstruction (C2) | C1, C2 | Textured targets; denser capture; fall-back to metric capture |
| Recognition gap without imagery | Dark, small, or thin components are ambiguous in a point cloud alone | S1: 9 of 16 monitors and 2 of 6 sockets found (C4); S2: 2 of 8 trays and 6 of 9 sprinklers (C4) | C4 | Add posed imagery (C5) or use a scan with photographs (C3) |
| Over-segmentation and plausible completion | Extra or duplicated components | C1 in S1: 8 columns modeled instead of 4; 53 false positives in S2 (C1) | C1, C1+P | Evidence-linked inventory; review of false positives |
| Inaccurate or over-trusted prior | Inaccurate parts of the prior propagate into the new model; own evidence under-used | S1: rough prior with inaccurate furniture and lights; light error 59 cm (C2+P) vs. 11 cm (C2); C2+P worse than C2 | C1+P, C2+P | State the reliability of each part of the prior; treat it as a hypothesis; change detection |
| Topology errors | Missing or wrong connections between fittings and pipes | S2 connection recall 7% (C1), 14% (C2) | C1, C2 | Topology-aware prompting and checks |
| Self-referential QA | Agent checks measure consistency with its own assumptions | All agents reported internal checks only | All | Independent, verified reference |

## 6.4 Recommended Configurations by Use Case

Table 17 translates the results into recommendations. They are derived from two scenes and one agent and should be read as indicative. In general, the smartphone LiDAR scan is the default for fast, low-cost modeling of rooms and their contents, scale-free imagery suffices for inventories, an accurate prior adds the building fabric at no capture cost, and TLS, preferably with posed panoramas, is reserved for tolerance-critical and large-scale work.

**Table 17** Configuration recommendations by use case.

| Use case | Required position accuracy | Components | Recommended configuration(s) | Rationale |
|---|---|---|---|---|
| Space planning and asset inventory | Decimeters | Furniture, equipment, counts | C2 or C3; C1 with an accurate prior | Recall 92–100% in S1 with 1–3 min of capture |
| Renovation design | ≤ 5–10 cm for fixed elements | Architecture, fixed elements | C3, or C1/C2 with an accurate prior | Wall and column offsets of 3–4 cm (C3); a prior fixes the fabric |
| As-built verification and tolerance-critical work | ≤ 2–5 cm | Fixed elements, MEP for prefabrication | C5 (or C4) | Median errors of 1–5 cm; highest geometric fidelity |
| MEP maintenance and coordination | ≤ 5 cm, correct systems and connections | Pipes, sprinklers, trays, fittings | C5; C3 as low-cost alternative | Recall 85–93% and connection recall 77–89%; scale-free inputs fail |
| Model update after layout change | Decimeters | Movable assemblies | C3; C2 for inventory only | Metric positions of movable contents; priors do not locate them |
| Digital-twin visualization | Decimeters | Broad, open-vocabulary | C2 or C3 | High semantic recall at low cost |

### 6.5 Limitations and Future Work

The study evaluates one agent and one LLM version with one run per configuration; agent capabilities change rapidly and agent outputs are non-deterministic, so the absolute results will shift with newer models and repeated runs, although the relative effect of the capture evidence is expected to be more stable. The reference models were derived from the C5 outputs and reviewed manually, so C5 is not independently evaluated and its geometric agreement is likely overstated. The experiments cover two scenes of limited size, so the observed differences between configurations, and the recommendations derived from them, require validation with repeated runs and more scenes before they can be generalized. The capture times of the panoramas were estimated rather than measured. The study also does not compare the agent with other automated Scan-to-BIM methods, such as 3D deep learning, hybrid 2D–3D, and vision-foundation-model-based methods, and it does not test preset component libraries or feed-forward reconstruction models as additional inputs [16,17]. Future work should address these points and extend the evaluation to larger, more numerous, and more complex scenes.

A particularly important next step is multi-agent collaboration with a standard operating procedure (SOP) for checking. In this study, the same agent built and checked each model, so its checks measured consistency with its own assumptions. A second agent, given only the evidence and the model, could act as an independent checker: one agent builds, the other verifies, and disagreements return to the builder as correction requests. The SOP would specify, for each component class, what has to be checked (presence, class, position, dimensions, and, for MEP, system and connections), how it is checked (for example, against point-cloud slices or by reprojection into images), and how much evidence is required, such as the minimum number of photographs from different viewpoints into which a component must be projected and agree before its position is accepted. Such a protocol would turn the agent's self-reports into auditable verification and would make the reliability of each component explicit.

# 7. CONCLUSIONS

This paper evaluated Agent2BIM, an agent-based Scan-to-BIM framework in which AI agents build models directly in 3D modeling software, across seven reality-capture configurations, from five 360° panoramas to terrestrial laser scanning, with and without a prior model, in a laboratory and an MEP scene, against reference models verified against TLS. Within the limits of two scenes and one run per configuration, four conclusions follow. (1) Recognition was not the main bottleneck: the agents detected 84–100% of the 103 laboratory components from every input, although detection was counted without a distance limit. Position was: the median error fell from 61 cm and 26 cm for panoramas and video to 8.8 cm for a smartphone LiDAR scan and 4.8 cm for TLS, and the share of components within 20 cm rose from 9% to 86% (RQ1). (2) A prior model supplied the frame that scale-free inputs lack and raised the MEP F1 score from 0.14 to 0.60 and from 0.51 to 0.78; a rough prior improved the architecture of the laboratory but, where its furniture and lights were inaccurate, introduced errors, so the value of a prior depends on its accuracy (RQ2). (3) Imagery and metric geometry are complementary: TLS without images missed small and dark components, posed panoramas recovered them, and the smartphone LiDAR scan, which records both, offered the most favorable balance of effort and accuracy in both scenes tested (2–3 min on site, USD 13 per run, 86% of laboratory components within 20 cm), while agent cost was similar for all inputs; scale-free imagery suffices only for inventory-grade models of furnished rooms, and TLS remains necessary for tolerance-critical work (RQ3). (4) The characteristic failures are scale and layout ambiguity, the inability to place elevated networks without depth, over-trusted priors, and self-referential checking, which call for metric anchors and independent verification (RQ4). The value of AI-driven Scan-to-BIM will therefore depend less on modeling effort than on capture strategies that supply the right evidence efficiently and on verification, for example by a separate checking agent, that makes the reliability of each component explicit.

## ACKNOWLEDGMENT

This research was partially supported by the Center for Sand Hazards and Opportunities for Resilience, Energy, and Sustainability (SHORES Research Center), funded by Tamkeen under the NYUAD Research Institute Award CG013. The authors also thank Asylkhan Kali and Ali Zeid for their contributions to the preliminary experimental trials.

## DATA AND CODE AVAILABILITY

The task prompts are given in Appendix A. The capture data, the reference models, the agent-generated models with their generating scripts and modeling summaries, and the evaluation scripts are available from the corresponding author upon reasonable request.

## APPENDIX A. TASK PROMPTS

The prompts below were used for the two scenes. The quoted input description is replaced according to the configuration: "360-degree images" (C1), "smartphone video" (C2), the exported smartphone LiDAR scan data (C3), "point cloud" (C4), and "point cloud, 360-degree images, and corresponding camera poses" (C5); in C1+P and C2+P, the existing IFC model of the space is added to the input description. For C1 and C2, the reference dimension was not included in the initial task prompt but was provided by the operator in response to a question raised by the agent immediately after receiving the task. The dimensions provided were: in S1, the width of the large double-door opening (about 1.70–1.80 m); in S2, the width of the glass door (about 1.9 m) for C1 and the length of the straight pipe segment above the door (about 0.95 m) for C2.

**Scene S1.** "Based on the 'point cloud, 360-degree images, and corresponding camera poses' provided in the files, please create an accurate model in Blender. The target components include walls, columns, floors, doors, tables, chairs, shelves, cabinets, whiteboards, ladders, monitors, robotic arms, mobile robots, quadruped robots (robot dogs), lighting fixtures, electrical outlets, fire extinguishers, other fire-safety equipment, and exit signs. Please ensure that the quantity, position, and dimensions of all components are accurate. Do not retain the original mesh in the final model."

**Scene S2.** "Based on the 'point cloud, 360-degree images, and corresponding camera poses' provided in the files, please create accurate MEP pipeline and architectural models in Blender. Use straight pipe segments wherever possible, and include fire sprinkler heads for the red fire-protection pipelines. Please ensure that the quantity, position, and dimensions of all components are accurate, and that their connections are modeled correctly. Do not retain the original mesh in the final model."

## APPENDIX B. SUPPLEMENTARY RESULT MAPS

Figs. B1–B4 show the S1 results per component category: for each configuration, the plan footprints of the reference components (gray) and of the matched model components (teal) are drawn together with the false positives (orange) and false negatives (purple), and a line joins the centers of each matched pair, so that its length is the position error. Figs. B5–B11 show, for each configuration in scene S2, (a) the reference network with the true and false positives, (b) the missed reference components, (c) the position offset of the true positives on a common color scale, and (d) the local connections at the fitting groups that were recovered or missed; the hatched areas lie outside the evaluated region.

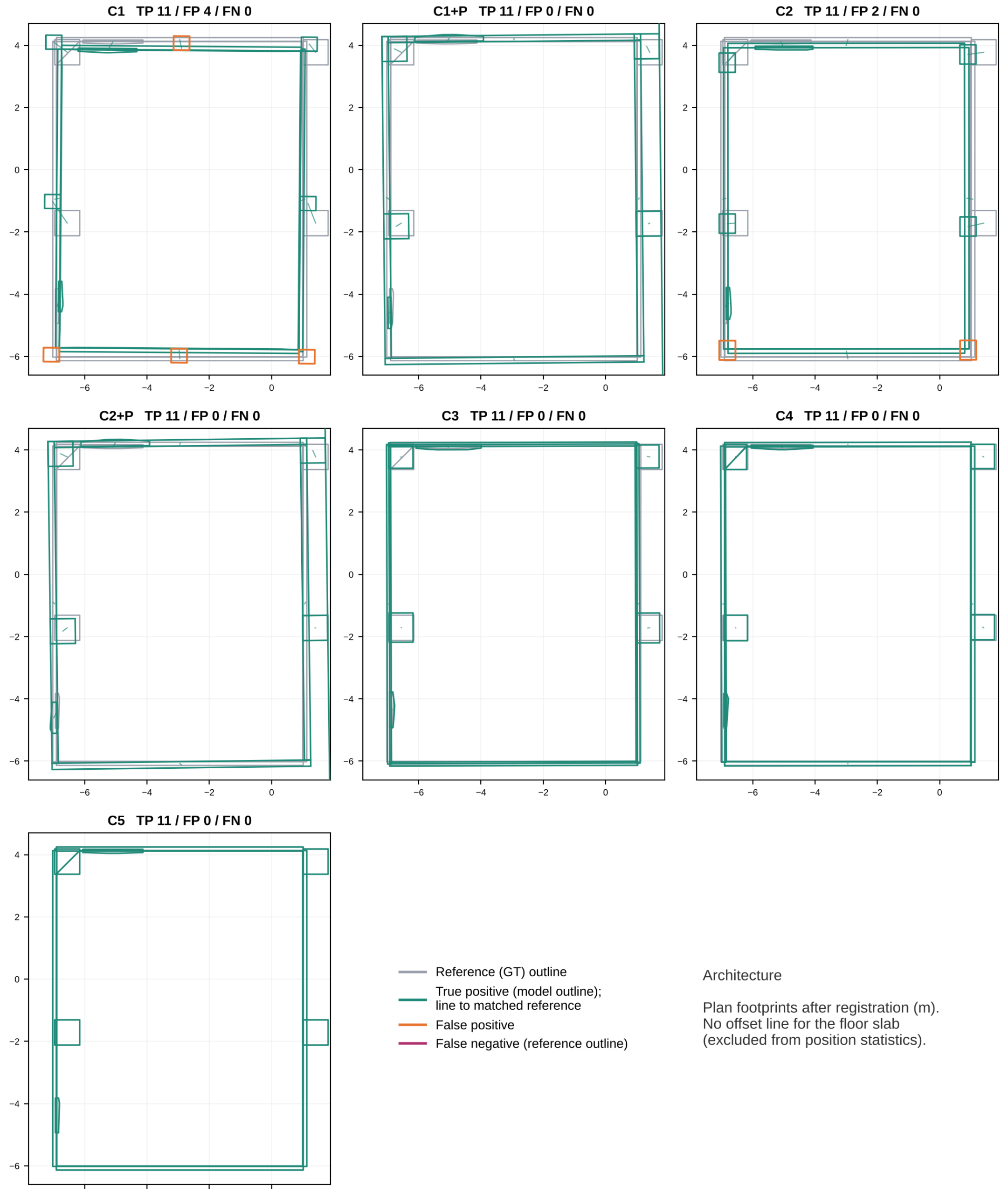

**Fig. B1** Scene S1, architecture: reference and model components of the seven configurations in plan.

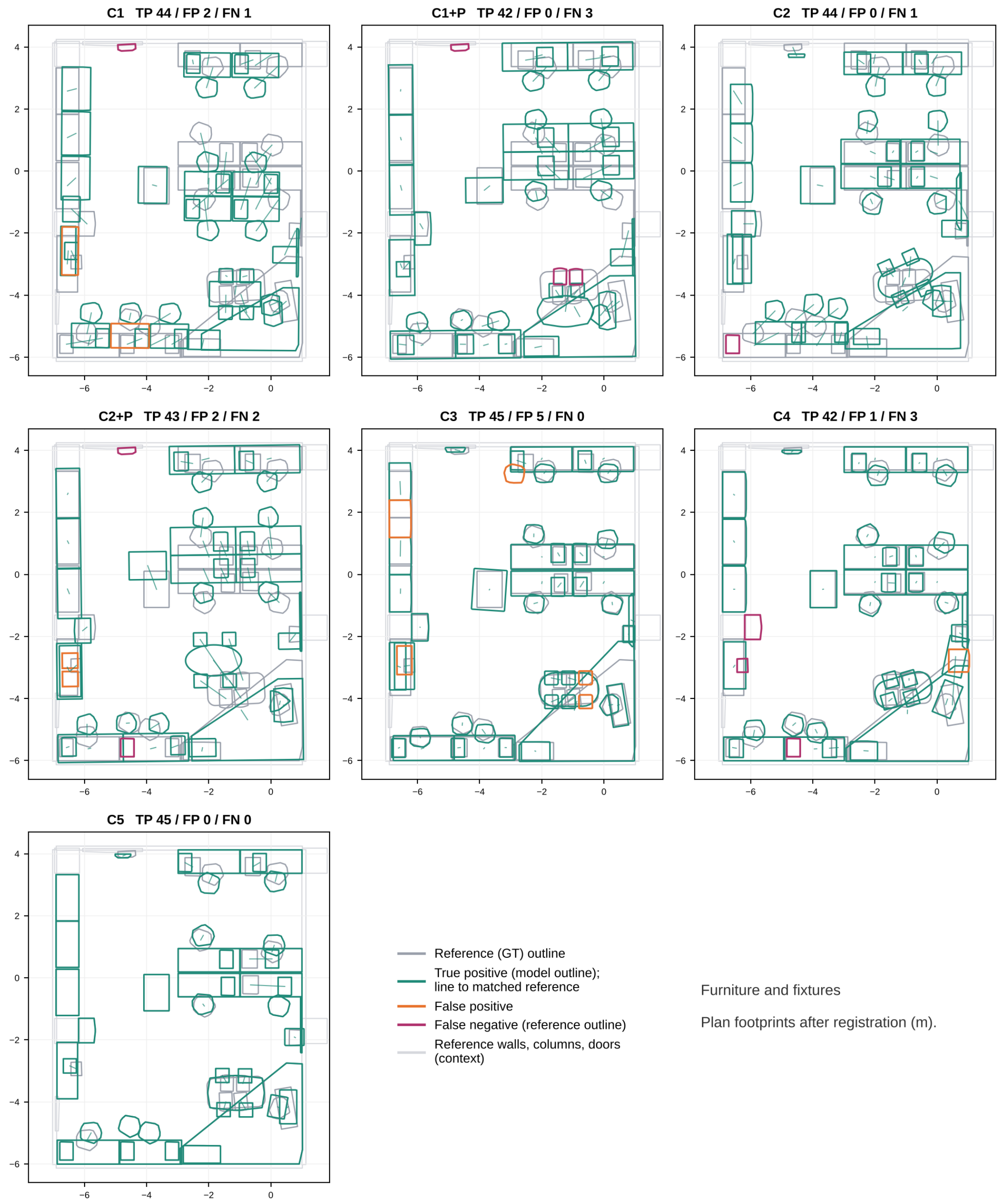

**Fig. B2** Scene S1, furniture and fixtures: reference and model components of the seven configurations in plan.

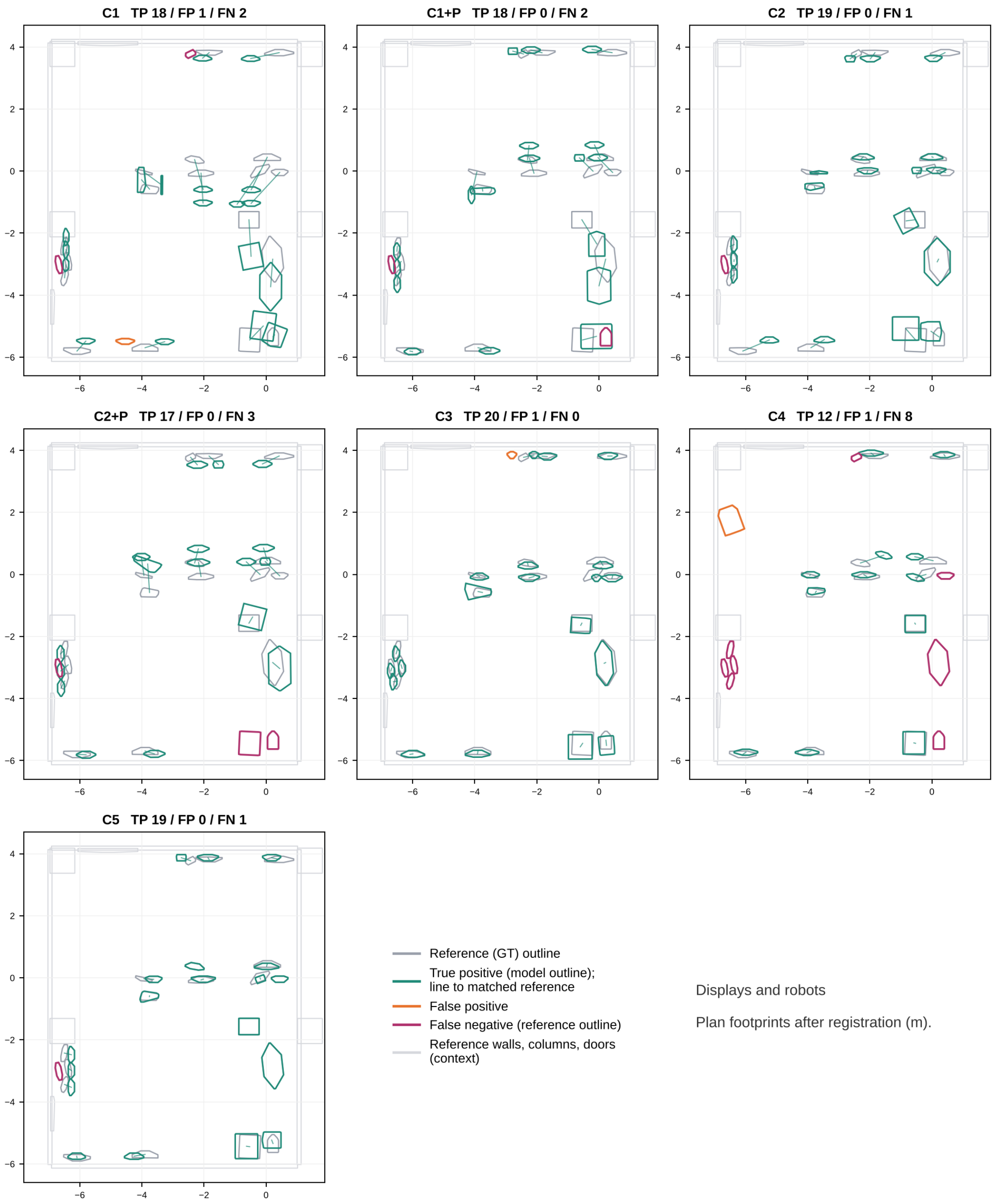

**Fig. B3** Scene S1, displays and robots: reference and model components of the seven configurations in plan.

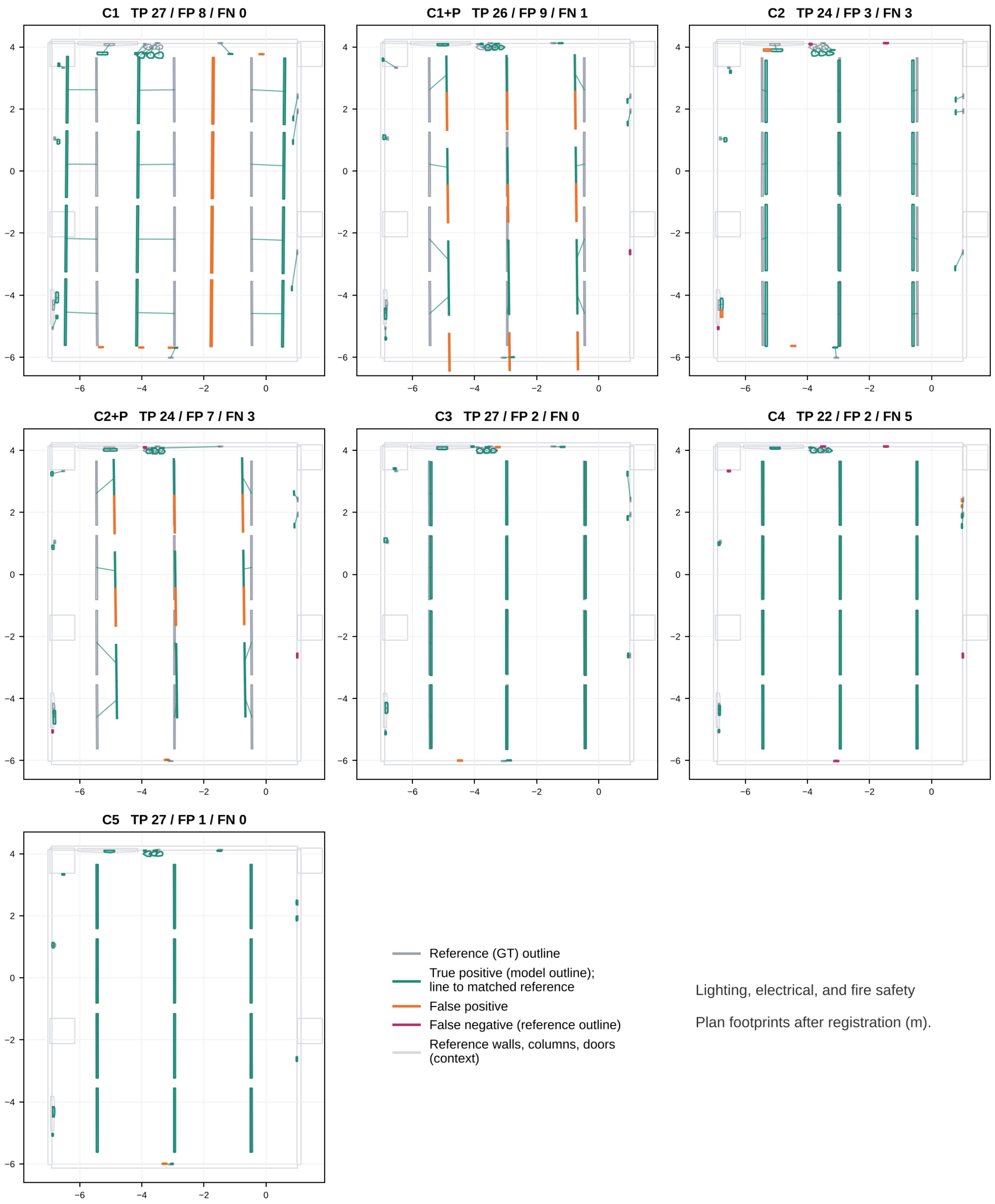

**Fig. B4** Scene S1, lighting, electrical, and fire-safety components: reference and model components of the seven configurations in plan.

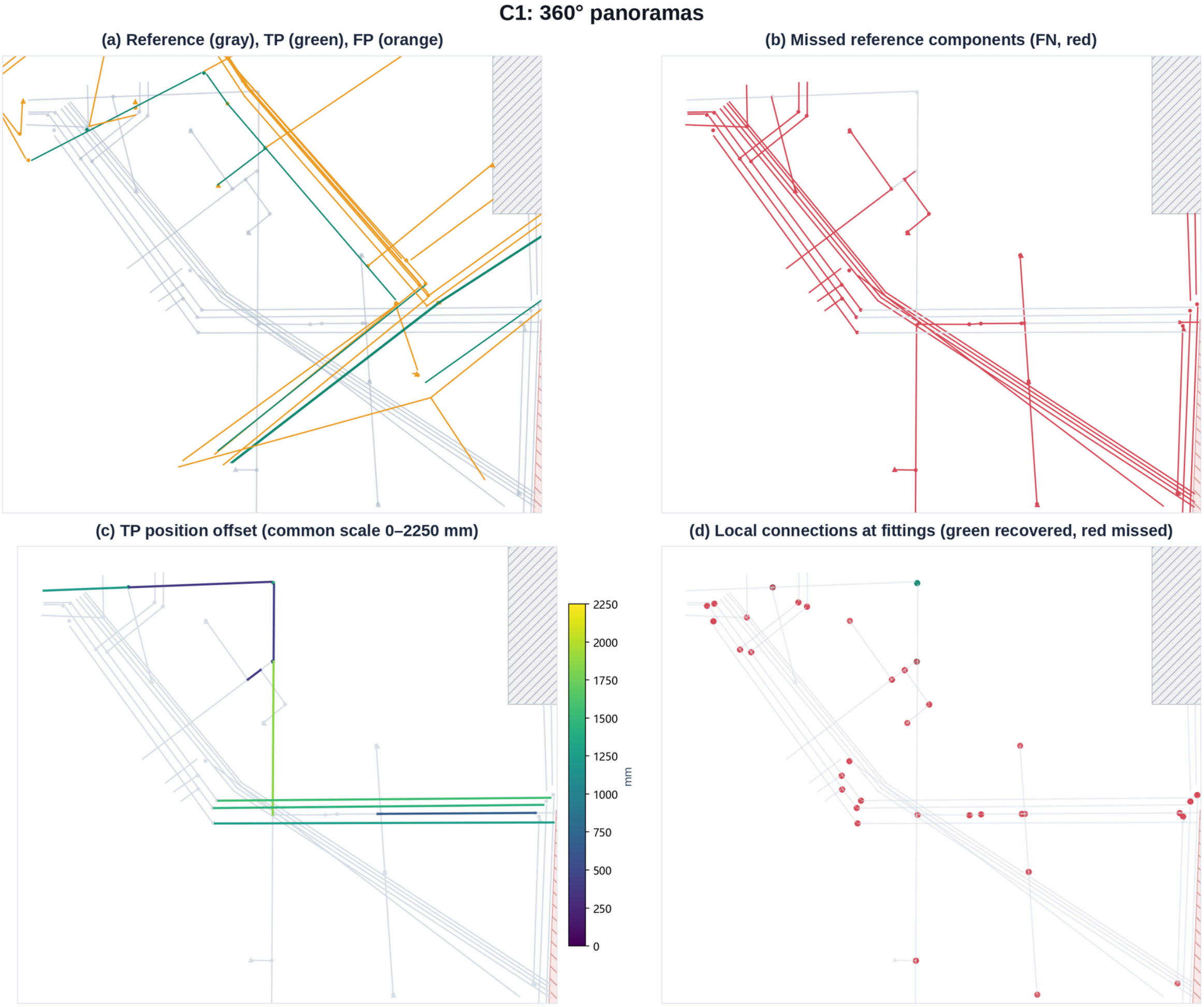

**Fig. B5** Result maps of C1 (360° panoramas) in scene S2.

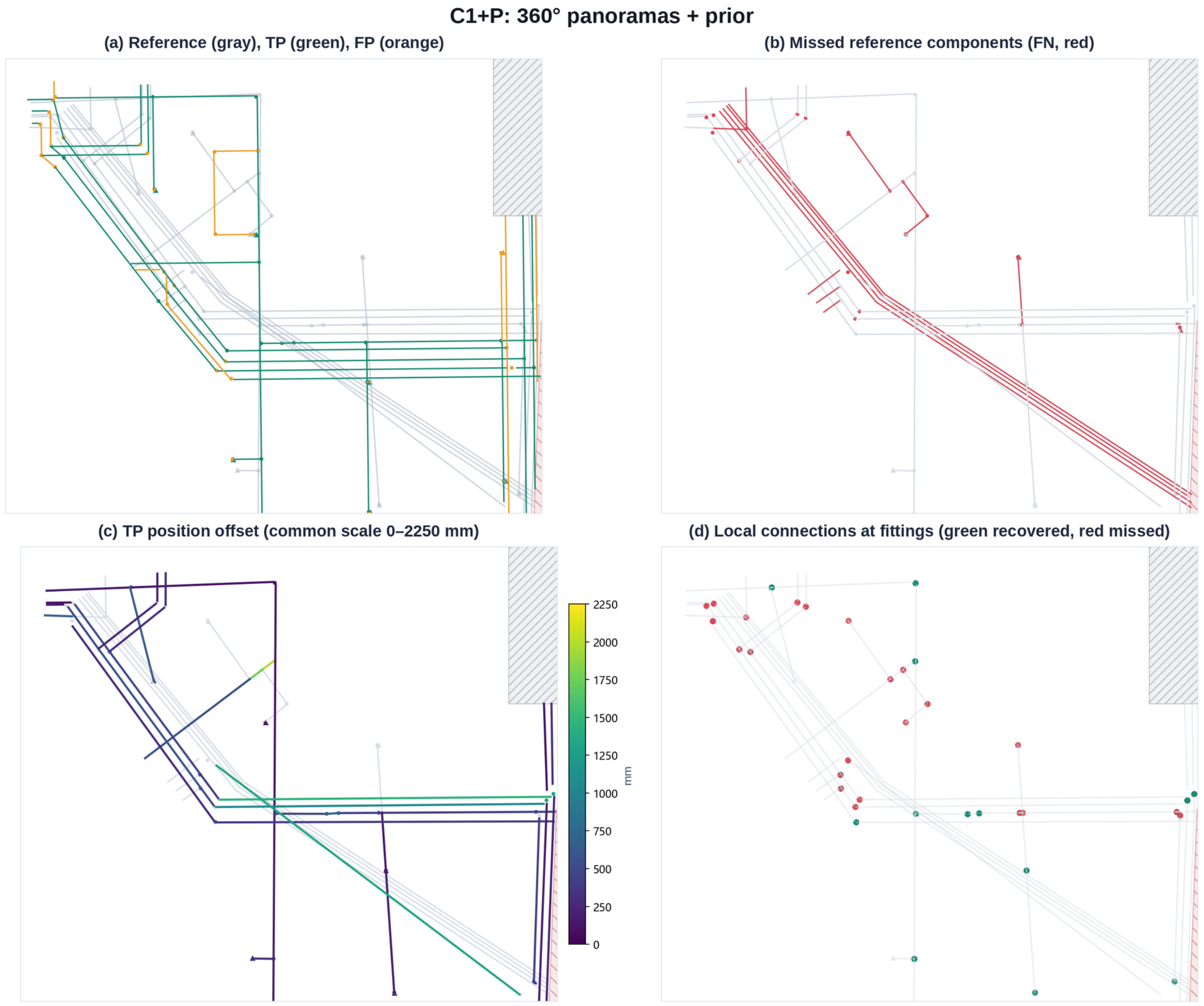

**Fig. B6** Result maps of C1+P (360° panoramas with prior) in scene S2.

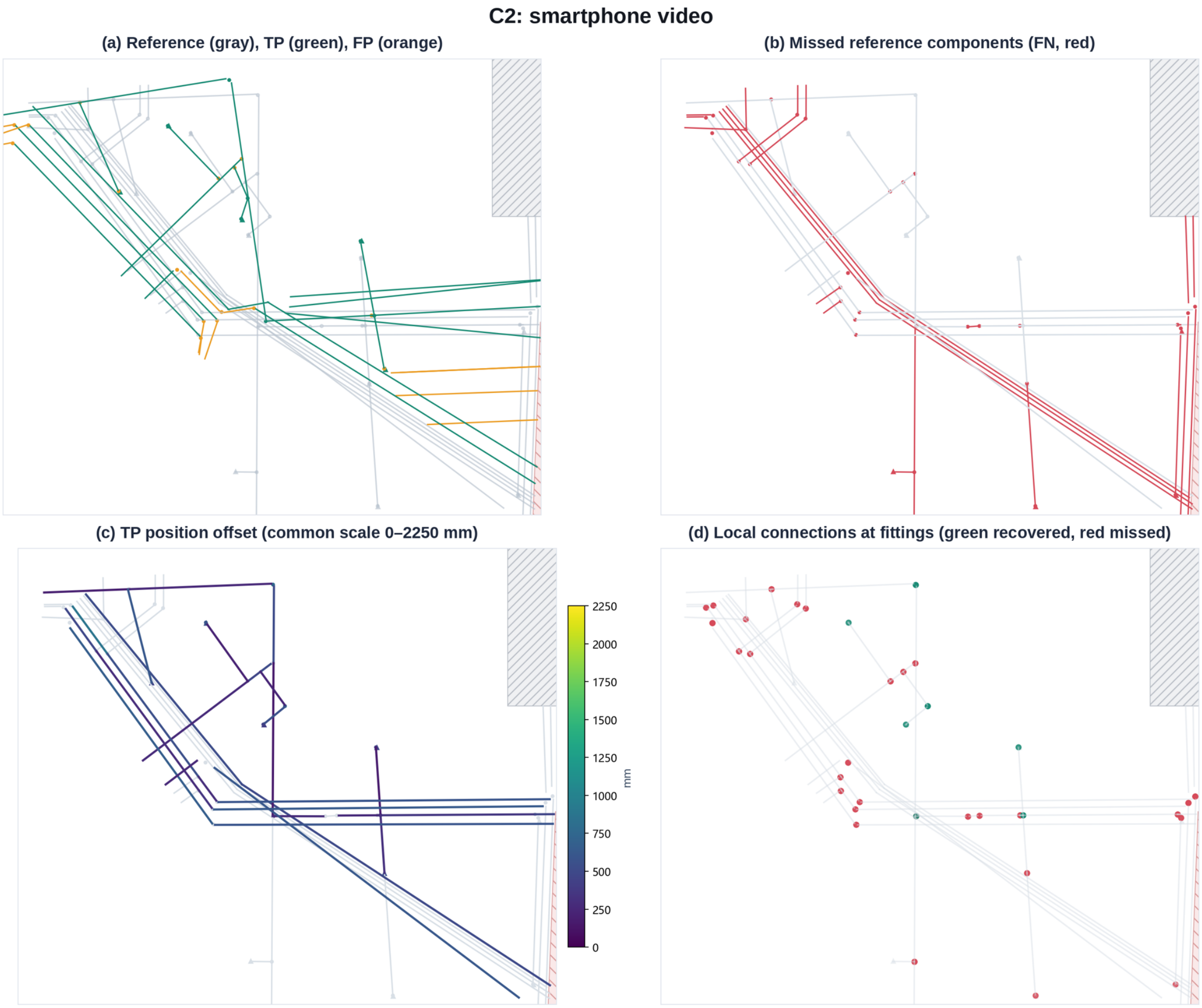

**Fig. B7** Result maps of C2 (smartphone video) in scene S2.

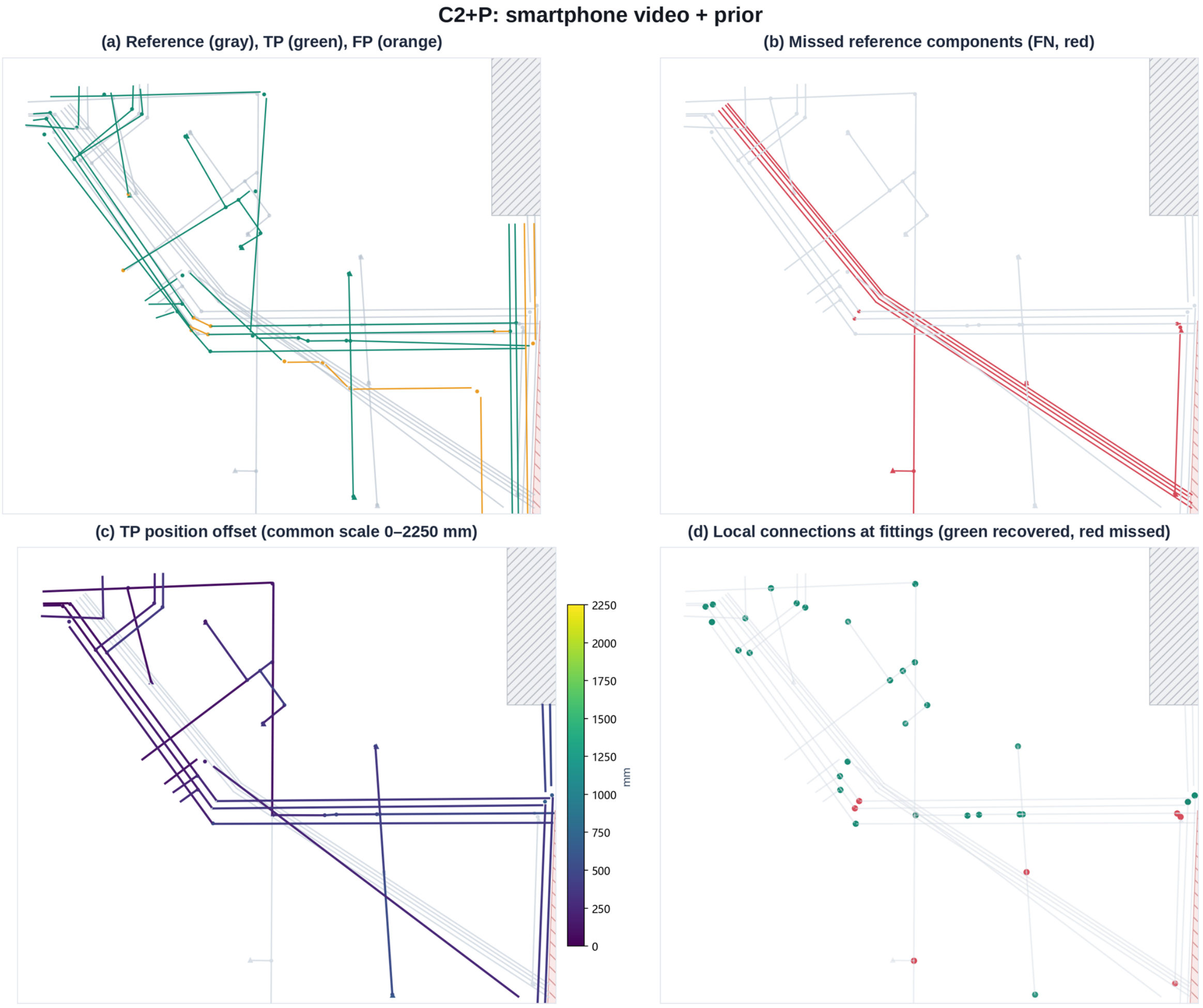

**Fig. B8** Result maps of C2+P (smartphone video with prior) in scene S2.

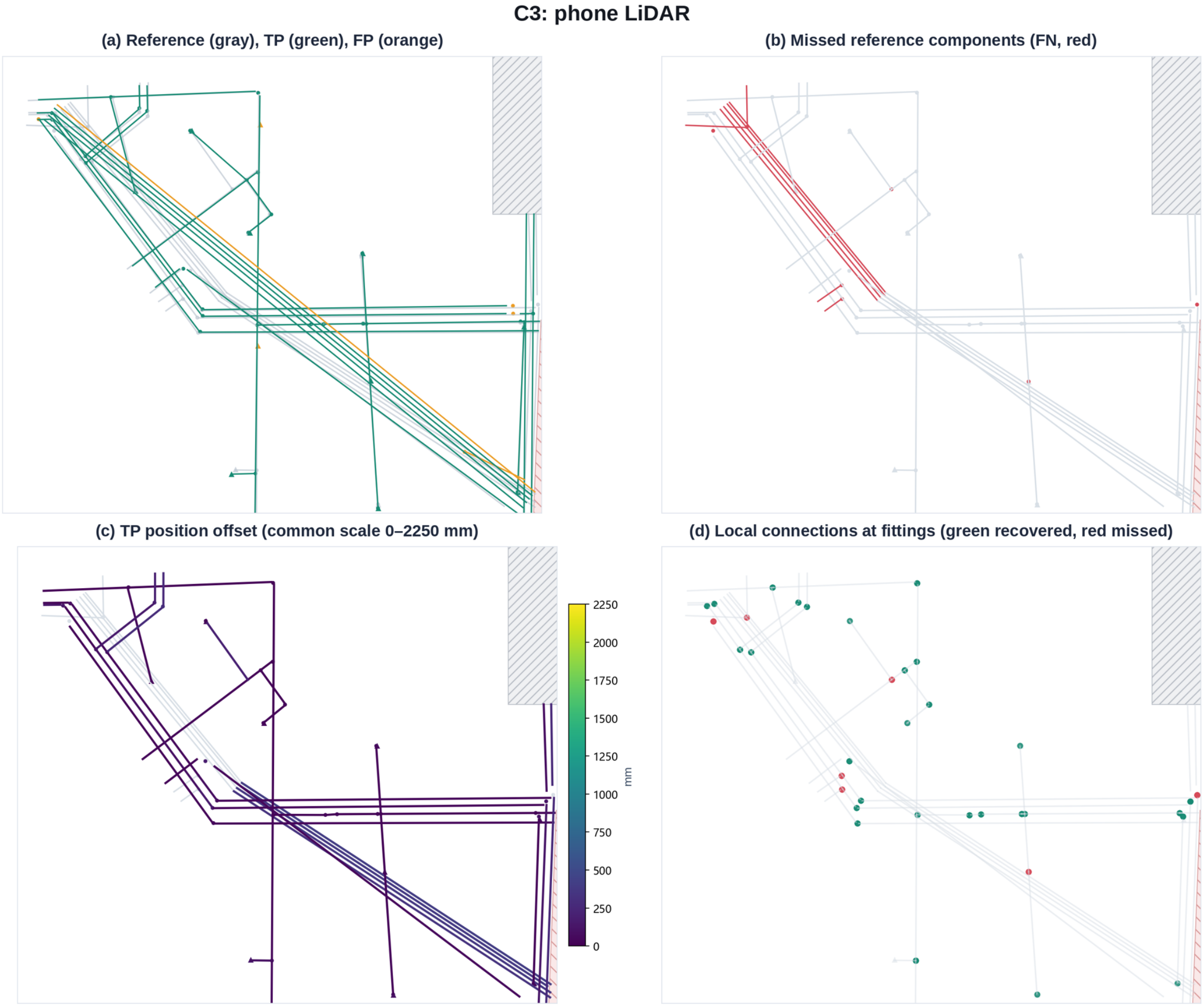

**Fig. B9** Result maps of C3 (smartphone LiDAR scan) in scene S2.

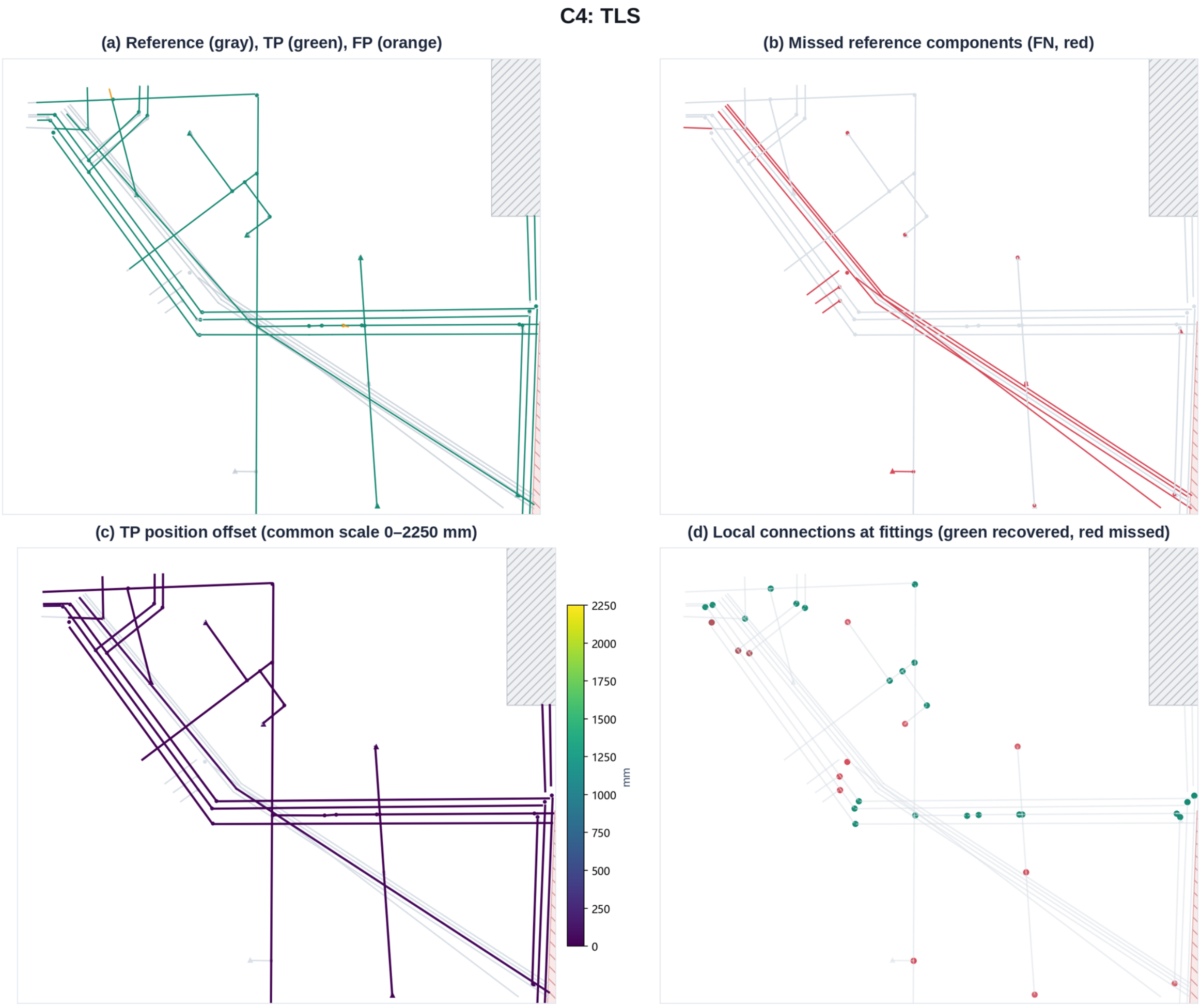

**Fig. B10** Result maps of C4 (TLS) in scene S2.

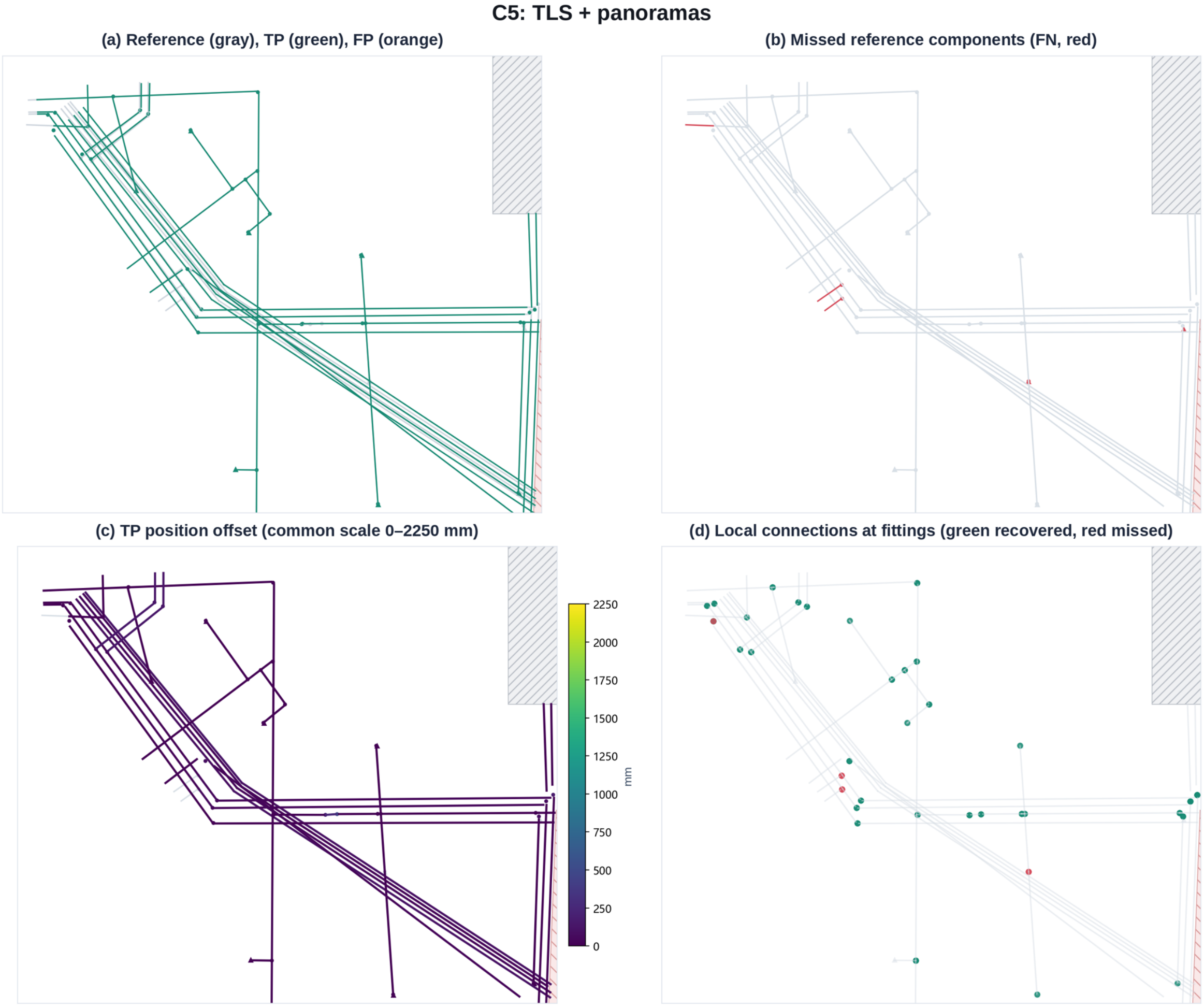

**Fig. B11** Result maps of C5 (TLS with posed panoramas) in scene S2.